\documentclass[mlmain]{jmlr}
\jmlrproceedings{PMLR}{Proceedings of Machine Learning Research}
\usepackage{longtable}% for long tables

\usepackage{booktabs}
\usepackage[load-configurations=version-1]{siunitx} % newer version
\providecommand{\yusu}[1]{}

\theorembodyfont{\upshape}
\theoremheaderfont{\scshape}
\theorempostheader{:}
\theoremsep{\newline}

\jmlrvolume{}
\firstpageno{1}
\editors{List of editors' names}

\jmlryear{2026}
\jmlrworkshop{Symmetry and Geometry in Neural Representations}

\title[Topological Invariance Simplification Discovers Behavioral Transitions]{Shaped by What's Missing: Topological Invariance Simplification Discovers Behavioral Transitions}

\author{%
\Name{Kaiwen Bian}$^{1,2}$ \Email{kbian@ucsd.edu}\\
\Name{Eric J. Leonardis}$^{2}$ \Email{eleonardis@salk.edu}\\
\Name{Yuanjia Yang}$^{1,2}$ \Email{scyang@salk.edu}\\
\Name{Charles Zhang}$^{3}$ \Email{charleszhang@fas.harvard.edu}\\
\Name{Yusu Wang}$^{1}$ \Email{yusuwang@ucsd.edu}\\
\Name{Talmo D. Pereira}$^{1,2}$ \Email{talmo@salk.edu}\\[0.5em]
\addr{%
$^{1}$University of California San Diego\\
$^{2}$Salk Institute for Biological Studies\\
$^{3}$Harvard University
}
}

\begin{document}

\maketitle

\begin{center}
\small
\textbf{Website:} \url{https://kbian.org/TopoMimic-website/} \\
\textbf{Code:} \url{https://github.com/KevinBian107/TopoMIMIC}
\end{center}

\begin{abstract}
Understanding how animals move between behaviors is a central question in neuroethology, yet most segmentation methods first partition time into discrete states, leaving transitions as the unmodeled residue between them. We take the opposite view: in a suitable feature space, sustained behaviors are dense regions of the trajectory's density landscape, and transitions are the sparse corridors and loops that connect them. Because this structure is a property of the landscape's shape rather than of any particular coordinates, it can be read off directly with topology and should persist across feature spaces. We build this landscape from frame-by-frame trajectories, extract its discrete-Morse graph, and keep two kinds of structure: paths, the corridors a trajectory follows between dense behavioral regions, and cycles, closed routes whose entry and exit take different ways. Clustering these structures yields transitions directly, with no upstream state partition. On a rat behavior dataset across three feature spaces, including the latent space of a neural-network controller that imitates the animal in a physics simulator, the approach recovers transitions at coverage comparable to state-space and topological-clustering baselines, while carrying substantially more information about the transition identity, forming clusters at least as tight as the baselines', and matching annotated transitions with less reliance on time-warping. More broadly, the results show that behavioral transitions are a topological feature of a representation: recoverable from the shape of its trajectories, and consistent across distinct feature spaces.
\vspace{0.2cm}

\end{abstract}

\section{Introduction}

How animals orchestrate transitions between behavioral modes (such as walking to rearing) is a central challenge in neuroethology, especially when the behavior is diverse. Transitions happen at many time scales, yet existing tools describe them only at macro resolution, because they treat a transition as \emph{downstream of a coarse-graining} applied to the raw trajectory. An autoregressive HMM or a switching linear dynamical system (SLDS)~\citep{wiltschko2015mapping,weinreb2024keypoint,hsu2021bsoid} assigns each frame a syllable and summarizes transitions as a matrix over syllables. A variational autoencoder~\citep{luxem2022identifying} partitions a latent embedding and treats transitions as switches between partitions. Mapper~\citep{singh2007topological,saggar2022precision} builds a cluster graph over overlapping density patches and reads transitions off its topology. The coarse-graining always happens first, so a transition is inherited from the state assignment rather than from the trajectory, and \emph{how} the body reconfigures frame by frame is lost before any analysis is run.

Deep reinforcement learning imitation policies expand this toolkit by emulating behavior in a physics simulator that reproduces recorded motion trajectories~\citep{merel2020deep,aldarondo2024virtual,mimicmjx2025}. In MIMIC-MJX~\citep{mimicmjx2025} (Figure~\ref{fig:mimic-pipeline}), inverse kinematics was performed on 3D pose recordings of freely behaving rats to produce joint angles (qpos) and serve as an imitation target. Then an encoder maps a short qpos reference window into a \emph{latent intention} $i_t$, and a decoder combines $i_t$ with proprioception to produce a biomechanically realistic action. Latent intentions and joint angles are control-relevant spaces that we utilize to study \emph{how} behaviors transition, in comparison with the egocentric positions of body parts (keypoints) for contrast.

We extract transition structure directly from the raw trajectory, applying discrete-Morse graph reconstruction~\citep{DeyWangWang2018,MageeWang2022,WangWangLi2015} to the density-weighted trajectory point cloud and recover the one-dimensional skeleton of its density terrain (Figure~\ref{fig:dmg-intuition}A shows the canonical road-network example). Two primitives live on this skeleton (Figure~\ref{fig:dmg-intuition}B): \emph{DM-paths}, the corridors between dense behavioral regions, and \emph{DM-cycles}, the loops left when a behavior's entry and exit arms occupy distinct routes. These are feature-level objects where one trajectory yields several primitives containing behavior segments. Our contributions are two fold: \textbf{(i)} We define these primitives and a Wasserstein clustering over them, a transition-centric representation needing no upstream state partition. \textbf{(ii)} We evaluate our method on a rat behavioral dataset in three feature spaces against Keypoint-MoSeq and Mapper baselines, showing its advantage lies in transition information, internal consistency, and agreement with ground truth behavior labels.

\section{Related Work}

Unsupervised discovery of behavioral structure from pose data has been approached through autoregressive HMMs (MoSeq)~\citep{wiltschko2015mapping}, discriminative classifiers on pose features (B-SOiD)~\citep{hsu2021bsoid}, variational autoencoders on temporal segments (VAME)~\citep{luxem2022identifying}, and keypoint-conditioned AR-HMM/SLDS (Keypoint-MoSeq) ~\citep{weinreb2024keypoint}; all segment trajectories into discrete syllables and treat transitions as instantaneous switches, describing them at the resolution set upstream by clustering or syllable assignment. Topology has also been increasingly applied to neuroscience, including persistent-homology analyses of neural manifolds~\citep{carlsson2009topology,gardner2022toroidal,rybakken2019decoding} and Mapper-based characterization of transition structure in neural dynamics~\citep{singh2007topological,saggar2022precision}. The discrete-Morse graph reconstruction we build on~\citep{forman1998morse,delgado2015skeletonization,gyulassy2007topologically,robins2011theory,sousbie2011persistent,WangWangLi2015,DeyWangWang2018,MageeWang2022} has itself been applied to segmenting neuroanatomical volumes~\citep{banerjee2020semantic}, and we give a more detailed account of it in Appendix~\ref{app:dm-details}; deep RL agents controlling simulated bodies have separately been analyzed as virtual model organisms~\citep{merel2020deep,aldarondo2024virtual}, but without examining topological structure. All of these analyze \emph{neural activity} manifolds, whereas we apply TDA to \emph{behavioral} manifolds, where topological features are transition routes rather than tuning curves; §\ref{sec:results} benchmarks both the segmentation and Mapper families against DM-primitives.

\section{Background}\label{sec:background}

\subsection{Animal behavior emulation in simulation}
In MIMIC-MJX~\citep{mimicmjx2025}, markerless motion capture is retargeted onto a biomechanical model in a physics simulator~\citep{todorov2012mujoco} and a reinforcement learning policy is trained to imitate the resulting kinematics (Figure~\ref{fig:mimic-pipeline}A). At each step the encoder maps a short joint angle reference window $s^{\text{ref}}_{t:t+5}$ to a latent intention $i_t$ under Gaussian and AR(1) regularization. The decoder then combines $i_t$ with the proprioceptive state $s_t$ to emit an action $a_t$ (Figure~\ref{fig:mimic-pipeline}B). Crucially $i_t$ represents \emph{latent action} rather than posture: it encodes the change the policy applies to reach the next reference joint angle pose.

We work in three representations at increasing levels of abstraction (details in Appendix~\ref{app:feature-spaces}): \textbf{egocentric keypoints (ekp, $\mathbb{R}^{60}$)}, the 20 keypoint 3-D positions in a body-fixed frame; \textbf{joint angles (qpos, $\mathbb{R}^{67}$)}, with the 7-dim global root pose removed; and \textbf{intention ($\mathbb{R}^{16}$)}, the learned latent, drift-invariant by construction. These describe, respectively, where the body is, how it is posed, and what change drives it toward the next pose.

\subsection{Discrete-Morse based graph skeletonization (DM-G)}\label{sec:setup}
Here we provide the intuition behind the discrete-Morse based graph skeletonization framework; Appendix~\ref{app:dm-details} gives the full account, illustrated in Figure~\ref{fig:dmg-intuition}A. We introduce it through the road-network reconstruction problem of~\citet{WangWangLi2015}, which is the setting the method was developed for and the easiest place to see what the construction recovers. There a density field derived from the locations of many cars concentrates around a hidden road network, but the data are noisy and the sampling non-uniform, so the network cannot simply be read off by thresholding (similar to the behavior case). Let $f: \mathbb{R}^d \to \mathbb{R}$ denote that density field. Viewing $f$ as a terrain, the hidden roads are its mountain ridges, captured formally by the \emph{1-stable manifolds} of $f$: the integral lines that connect local maxima to saddles. These curves connect density peaks through saddle points, separating low-density basins, and their union forms a graph that we take as the hidden skeleton.

These 1-stable manifolds are differential objects and can be hard and noisy to estimate in the discrete setting, especially in higher dimensional spaces. Following~\citet{DeyWangWang2018}, one instead works with discrete Morse theory~\citep{forman1998morse}, a combinatorial analog of Morse theory over cell (and therefore simplicial) complexes, which admits a simple combinatorial strategy for reducing the topological complexity of a discrete Morse function and thereby simplifying the resulting 1-stable manifolds. Persistent homology~\citep{edelsbrunner2010computational,dey2022computational} is used to guide that simplification, so that only ``important'' topological features are preserved. We build on this formulation and on its extension to high-dimensional point-cloud data~\citep{MageeWang2022}.

\section{Methods}
\begin{figure}[t]
    \centering
    \includegraphics[width=\textwidth]{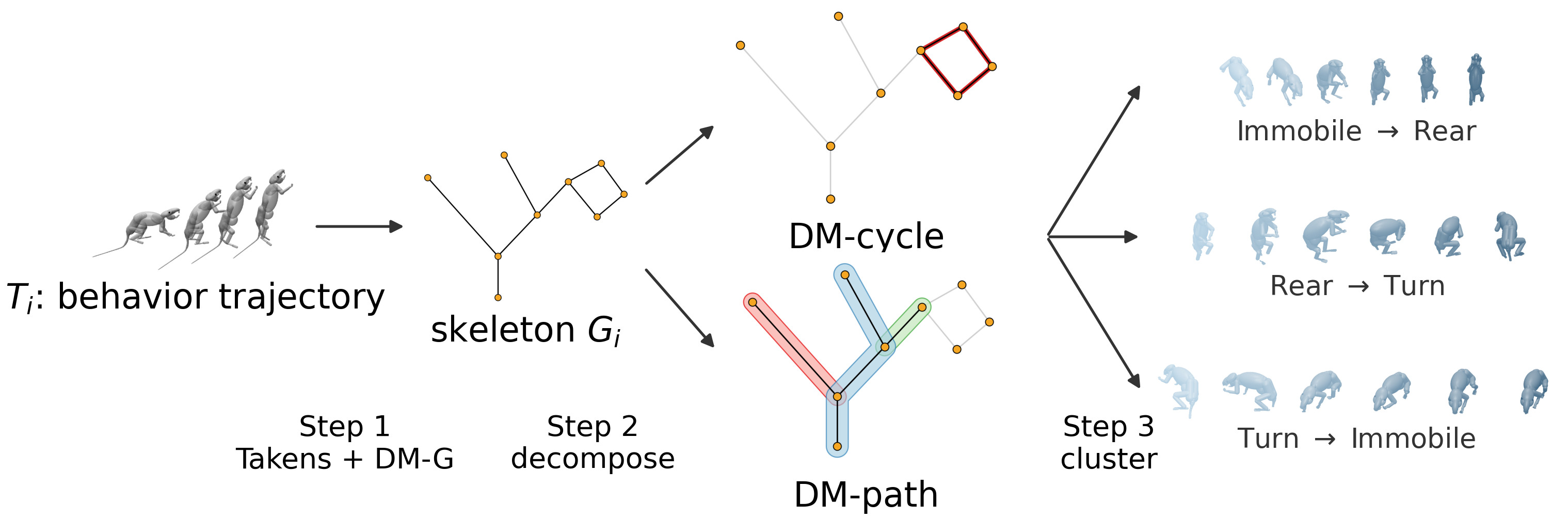}
    \caption{\textbf{Method overview.} A trajectory $T_i$ is Takens-embedded and skeletonized into a discrete-Morse graph $G_i$ (Step 1), decomposed into DM-cycles ($H_1$ loops) and DM-paths (the split tree) (Step 2), and the pooled primitives are clustered under $W_2$ into behavioral transition clusters (Step 3). Arrows mark each step's output as the next step's input. The two primitives are illustrated on a synthetic skeleton in Appendix Figure~\ref{fig:dmg-intuition}.}
    \label{fig:pipeline}
\end{figure}

Our input is a collection of recorded rat behavior trajectories, each a window of markerless motion-capture frames retargeted onto a biomechanical model in a physics simulator (§\ref{sec:setup}). The encoder $E$ of the MIMIC-MJX imitation policy~\citep{mimicmjx2025} maps each to a time series $T := \{p_1, \ldots, p_n\}$ in the latent space $\mathcal{Z} = \mathbb{R}^D$. The same pipeline also runs unchanged on the joint-angle (qpos) and egocentric-keypoint (ekp) representations, which instantiate $\mathcal{Z}$ at other dimensions. Three steps follow on the collection $\mathcal{T} = \{T_1, \ldots, T_M\}$ (Figure~\ref{fig:pipeline}): skeletonize each $T_i$ into a DM-graph $G_i$ (§\ref{subsec:DMgraph}), decompose $G_i$ into topological primitives (§\ref{subsec:primitives}), and cluster the pooled primitives under a Wasserstein metric (§\ref{sec:cluster}).

\subsection{(Step 1) Discrete-Morse based skeletonization}
\label{subsec:DMgraph}

Consider a trajectory $T = \{p_1, \ldots, p_n\}$, with $p_i \in \mathcal{Z} = \mathbb{R}^D$ the feature vector at frame $i$. A single frame is an ambiguous descriptor of intention, as the same instantaneous configuration can occur within different movements. We therefore first apply a time-delay (Takens) embedding~\citep{takens1981detecting} so that each point carries a short window of recent motion: we map $T$ to the point cloud
\(
X = \{\mathbf{x}_1,\ldots,\mathbf{x}_{n'}\} \subset \mathbb{R}^{mD}, \mathbf{x}_t = [\, p_t,\, p_{t-\tau},\, \ldots,\, p_{t-(m-1)\tau} \,],
\)
a set of $n' = n-(m-1)\tau$ points, one per admissible frame, with the delay $\tau$ and embedding dimension $m$ fixed by the average-mutual-information and False-Nearest-Neighbors heuristics (qpos and ekp are first reduced to their leading principal components; Appendices~\ref{app:feature-spaces},~\ref{app:dmg-hp}). Each point of $X$ is therefore the short-time \emph{pattern} of how the signal evolves rather than a raw posture, position or action at a single instant.

Retaining the temporal index only so that skeleton vertices can later be mapped back to frames, we reconstruct the density-weighted graph skeleton of $X$ with the discrete-Morse based reconstruction algorithm of~\citet{MageeWang2022}, which extends~\citet{DeyWangWang2018} to high dimensions. It builds a sparse weighted Rips filtration over $X$ from a $k$-nearest-neighbor density estimate, writing $\rho$ for the induced inverse-density score (large where the local $k$-NN density is low), and then performs the persistence-guided discrete-Morse simplification of §\ref{sec:setup} at a persistence threshold $\delta$. The output is a one-dimensional graph skeleton $G_T$ whose vertices are a subset of the points of $X$ and whose edges trace the density ridges of the trajectory: a clip of $n \approx 500$ frames becomes a skeleton of a few hundred vertices, which Step 2 cuts into primitives of ${\sim}60$--$190$ frames each (Appendix~\ref{app:reduction}).

\subsection{(Step 2) Topological primitive extraction}
\label{subsec:primitives}

We now decompose the DM-skeleton $G_T$ into more refined \emph{topological primitives} which potentially capture clean individual behavior transitions.

\paragraph{DM-cycles.} Loops in $G_T$ mark a \emph{recurring movement pattern}: the embedded trajectory leaves a region and returns by a different route. Because each embedded point is a short motion window (the Takens delay embedding), such of embedding is a recurrence of \emph{dynamics} rather than returning to a single static pose. This is the same as a pendulum traces a line in position but a loop in phase space. Cycles flag the asymmetry between the outgoing and returning routes, so they are most informative on transitions with a large change in dynamics (Appendix~\ref{app:asymmetry}).

We take as DM-cycles a minimum cycle basis of $G_T$~\citep{hagberg2008networkx}: writing $len(\gamma)$ for the total edge weight of a cycle $\gamma \subseteq G_T$, the set $\Gamma^* = \{\gamma_1^*, \ldots, \gamma_g^*\}$ of $g = \beta_1(G_T)$ independent loops whose length sequence $\ell_1 = len(\gamma_1^*) \le \cdots \le \ell_g = len(\gamma_g^*)$ is lexicographically smallest. We instantiate $len(e)$ for $e = (u, v)$ as $\max(\rho(u), \rho(v))$, the inverse density at the sparser endpoint of $e$, rather than as Euclidean distance. Such instantiation makes its value in the lower-star filtration of $\rho$ on $G_T$ (not its sparse-Rips filtration value), so the basis prefers loops through denser regions and serve as a practical surrogate for the persistence-optimal basis of~\citet{MageeWang2022}. The basis is canonical, keeping cycles comparable across clips. A loopless skeleton contributes none and cycles share no edges with the DM-paths below since every basis cycle is closed.

\paragraph{DM-paths.} Removing $\Gamma^*$ from $G_T$, that is $G_T \setminus \Gamma^*$, leaves a forest $\mathcal{F} = \{F_1, \ldots, F_s\}$, $s \ge 0$. Nodes of each tree $F$ are peaks and saddles, so ordering them by the height function $h = -\rho: F \to \mathbb{R}$ makes $(F,h)$ a split tree and orders vertices as the density $f$ of Appendix~\ref{app:dm-details}. Such tree is also rooted at the lowest value with all non-leaf nodes upward forks (Figure~\ref{fig:dmg-intuition}B). Running the persistence algorithm on the filtration of $F$ induced by $h$, via a union-find data structure~\citep{dey2022computational}, yields saddle-maximum pairs $(s, m)$. The unique tree path between $s$ and $m$ is then one \emph{DM-path}. We further take the path from the global minimum to the global maximum (intuitively a persistent pair in the extended persistence diagram). We emphasise that this is not a new filtration. It is exactly the standard $0$-dimensional super-level-set persistence of a height function on a tree, equivalent under $h \mapsto -h$ to the sub-level-set formulation. What is specific to our construction is not the filtration but its reading. Each pair $(s,m)$ is taken as a transition corridor of the trajectory that is edge-disjoint from the DM-cycles. We keep every such pair, with no persistence pruning of the paths (sensitivity in Appendix~\ref{app:path-persistence}). Repeating over $\mathcal{F}$ converts each trajectory $T$ into a collection of DM-cycles and DM-paths, its behavior segments, each mapped back to its frames and split at temporal gaps (Appendix~\ref{app:dmg-hp}).

\paragraph{Why drift-invariant spaces.} Both primitives rest on a skeleton that forms only when behaviorally similar frames recur near each other, which fails under uncorrected drift. When the same behavior at different times drifts apart, no dense region recurs and neither loops nor corridors survive. Hence the spaces of §\ref{sec:setup}, and our expectation that primitives are cleanest in intention and qpos and weakest in ekp, the most drift-laden space.

\subsection{(Step 3) Wasserstein Clustering}\label{sec:cluster}
Step 2 yields a pooled collection $\mathcal{P} = \{\pi_1, \ldots, \pi_N\}$ over $\mathcal{T}$, each $\pi_j$ an ordered sequence of points in $\mathcal{Z} = \mathbb{R}^D$. The two primitive types and the three feature spaces are clustered separately, producing the DMC and DMP variants of §\ref{sec:results}. Primitives differ in length, so we read each as an empirical distribution $\mu_j = \tfrac{1}{|\pi_j|}\sum_{p \in \pi_j} \delta_p$ and compare them under the $2$-Wasserstein distance \citep{peyre2019computational}, which matches the regions of $\mathcal{Z}$ two primitives occupy instead of imposing a frame-by-frame correspondence, and so absorbs the local time-reparameterization between two instances of one behavior. Resampling every primitive to a common length $L = 100$ and solving the exact discrete optimal transport for each pair gives an $N \times N$ distance matrix $\mathbf{D}$, which we embed with metric MDS at a dimension selected per variant by within-cluster consistency (Appendix~\ref{app:mds-sweep}) before running $K$-means on the embedded coordinates. We call this Discrete-Morse Wasserstein $K$-means Clustering (DM-WKC) and apply the identical $W_2 \rightarrow \text{MDS} \rightarrow K\text{-means}$ backend to the baseline primitives of §\ref{sec:baselines}. Appendix~\ref{app:wkc} gives the construction in full.

\subsection{Baselines and Evaluation Metrics}\label{sec:baselines}\label{sec:metrics}

We compare against two baseline families (hyperparameters in Appendix~\ref{app:baseline-hp}). \emph{State-space:} Keypoint-MoSeq~\citep{weinreb2024keypoint} on intentions (KPMS-Int) and on egocentric keypoints (KPMS-KP). \emph{Topological clustering:} Mapper~\citep{singh2007topological} (Appendix~\ref{app:mapper-hp}) on all three spaces, using the Fiedler vector of the $k$-nearest-neighbour graph Laplacian as its 1D lens, a standard Mapper filter chosen independently of our own construction. Both families assign every frame a discrete label, a syllable or a graph node, so we build their transition primitives natively rather than through the $W_2$ backend of §\ref{sec:cluster}: each cluster is one ordered pair of consecutive sustained labels, and the most frequent such pairs are retained. We keep each method's own transition unit instead of imposing a common backend, and Appendix~\ref{app:information} corrects for the label-cardinality asymmetry this creates. We also include a \emph{full-clip} control (Full) that treats each clip's whole intention trajectory as one primitive, ablating the decomposition step itself.
Ground-truth labels (Appendix~\ref{app:labeling}) are used only for evaluation, never during clustering. All alignment metrics operate in the same root-stripped qpos space: \emph{dynamic time warping} (DTW) measures frame-level boundary alignment, and \emph{kernel inception distance} (KID)~\citep{binkowski2018demystifying} compares the frame distributions of a cluster and its ground-truth window. KID is a maximum mean discrepancy under the polynomial kernel $k(x,y) = ((x{\cdot}y)/d + 1)^3$; we borrow it from the generative-modelling literature, where it scores sample quality, and apply the kernel directly to our features. \emph{Internal consistency} is the mean pairwise DTW within a cluster, and we decompose the warping path into a sync ratio and a non-diagonal fraction (Table~\ref{tab:main}, Appendix~\ref{app:tables}). Beyond alignment we evaluate each coarse-graining via the per-frame mutual information $I(T;Y)$ between its label $T_t$ and the directed-transition class $Y_t$, with cardinality-corrected AMI~\citep{vinh2010information} (Appendix~\ref{app:information}).

\section{Results}\label{sec:results}

All methods are evaluated on the same rat dataset against the auto-labeling ground truth of Appendix~\ref{app:labeling}. Twelve variants are compared. Six of them are ours, the DM-cycle (DMC) and DM-path (DMP) primitives of §\ref{sec:cluster} on each of intention, qpos and ekp; the baselines are Mapper on the same three spaces and the two Keypoint-MoSeq streams KPMS-Int and KPMS-KP, together with the full-clip control. Ours are marked $^*$ in Table~\ref{tab:main} and in the per-variant tables of Appendix~\ref{app:tables}. We report alignment and coherence (§\ref{sec:alignment}) and transition information (§\ref{sec:information}), with where the primitives appear and what the clusters look like in Appendices~\ref{app:primitive-checks} and~\ref{app:cluster-examples}. Control experiments are collected in Appendix~\ref{app:controls}.

\begin{table}[!htbp]
    \centering
    \caption{\textbf{Alignment, coherence, warping and distributional fidelity.} One fixed variant per family: DMC-Int, DMP-Qpos, Mapper-Qpos, KPMS-KP. Rows are directed transitions with turn direction ([L] left, [R] right, [S] straight). Lower is better in all four blocks. \emph{---} = not discovered; $^\dagger$ = one seed only. Summary rows give the per-column mean and std of the 3-seed $\pm$. $^*$Ours.}
    \label{tab:main}
\scriptsize
\setlength{\tabcolsep}{3pt}
\renewcommand{\arraystretch}{1.05}
\resizebox{\textwidth}{!}{%
\begin{tabular}{l|cccc|cccc|cccc|cccc}
\toprule
 & \multicolumn{4}{c|}{\textbf{DTW to GT ($\downarrow$)}} & \multicolumn{4}{c|}{\textbf{Consistency ($\downarrow$)}} & \multicolumn{4}{c|}{\textbf{Non-diag ($\downarrow$)}} & \multicolumn{4}{c}{\textbf{KID ($\downarrow$)}} \\
Transition & \shortstack{DMC$^*$\\{\tiny (Int)}} & \shortstack{DMP$^*$\\{\tiny (Qpos)}} & \shortstack{Mapper\\{\tiny (Qpos)}} & \shortstack{KPMS\\{\tiny (KP)}} & \shortstack{DMC$^*$\\{\tiny (Int)}} & \shortstack{DMP$^*$\\{\tiny (Qpos)}} & \shortstack{Mapper\\{\tiny (Qpos)}} & \shortstack{KPMS\\{\tiny (KP)}} & \shortstack{DMC$^*$\\{\tiny (Int)}} & \shortstack{DMP$^*$\\{\tiny (Qpos)}} & \shortstack{Mapper\\{\tiny (Qpos)}} & \shortstack{KPMS\\{\tiny (KP)}} & \shortstack{DMC$^*$\\{\tiny (Int)}} & \shortstack{DMP$^*$\\{\tiny (Qpos)}} & \shortstack{Mapper\\{\tiny (Qpos)}} & \shortstack{KPMS\\{\tiny (KP)}} \\
\midrule
Immobile$\to$Rear [L] & --- & \underline{0.854} & --- & \textbf{0.831} & --- & \textbf{0.803} & --- & \underline{0.821} & --- & \textbf{0.128} & --- & \underline{0.245} & --- & \textbf{0.002} & --- & \underline{0.006} \\
Rear$\to$Immobile [L] & --- & \underline{0.855$^{\dagger}$} & 0.879 & \textbf{0.855} & --- & \textbf{0.659$^{\dagger}$} & 0.743 & \underline{0.708} & --- & \textbf{0.111$^{\dagger}$} & 0.178 & \underline{0.165} & --- & 0.026$^{\dagger}$ & \textbf{0.009} & \underline{0.010} \\
Immobile$\to$Rear [R] & --- & \underline{0.880$^{\dagger}$} & --- & \textbf{0.848$^{\dagger}$} & --- & \textbf{0.718$^{\dagger}$} & --- & \underline{0.854$^{\dagger}$} & --- & \textbf{0.170$^{\dagger}$} & --- & \underline{0.173$^{\dagger}$} & --- & \underline{0.033$^{\dagger}$} & --- & \textbf{0.010$^{\dagger}$} \\
Immobile$\to$Rear [S] & 0.934 & \textbf{0.892} & \underline{0.895} & 0.967 & 0.766 & \textbf{0.632} & \underline{0.714} & 0.936 & 0.080 & \underline{0.063} & \textbf{0.055} & 0.096 & 0.032 & 0.024 & \textbf{0.010} & \underline{0.011} \\
Rear$\to$Immobile [S] & \underline{0.863$^{\dagger}$} & --- & \textbf{0.856} & 0.877 & \underline{0.802$^{\dagger}$} & --- & \textbf{0.765} & 0.816 & \underline{0.080$^{\dagger}$} & --- & \textbf{0.052} & 0.109 & 0.011$^{\dagger}$ & --- & \textbf{0.008} & \underline{0.008} \\
Immobile$\to$Turn [L] & 0.809 & \underline{0.788} & 0.801 & \textbf{0.770} & \textbf{0.652} & 0.724 & 0.690 & \underline{0.686} & \textbf{0.012} & \underline{0.019} & 0.033 & 0.036 & 0.009 & \underline{0.003} & 0.006 & \textbf{0.002} \\
Turn$\to$Immobile [L] & \underline{0.740} & \textbf{0.721} & 0.790 & 0.849$^{\dagger}$ & 0.601 & \textbf{0.585} & \underline{0.590} & 0.659$^{\dagger}$ & \textbf{0.032} & \underline{0.054} & 0.062 & 0.086$^{\dagger}$ & \textbf{0.005} & \underline{0.008} & 0.010 & 0.013$^{\dagger}$ \\
Immobile$\to$Turn [R] & 0.759 & 0.756$^{\dagger}$ & \textbf{0.715$^{\dagger}$} & \underline{0.747} & 0.703 & 0.729$^{\dagger}$ & \textbf{0.669$^{\dagger}$} & \underline{0.688} & \underline{0.013} & \textbf{0.005$^{\dagger}$} & 0.019$^{\dagger}$ & 0.025 & \underline{0.007} & 0.009$^{\dagger}$ & 0.008$^{\dagger}$ & \textbf{0.003} \\
Immobile$\to$Walk [L] & \textbf{0.759$^{\dagger}$} & --- & --- & --- & \textbf{0.928$^{\dagger}$} & --- & --- & --- & \textbf{0.080$^{\dagger}$} & --- & --- & --- & \textbf{0.003$^{\dagger}$} & --- & --- & --- \\
Rear$\to$Turn [L] & 0.826 & \textbf{0.778} & \underline{0.818} & 0.918 & 0.782 & \textbf{0.607} & \underline{0.643} & 0.795 & \textbf{0.142} & 0.169 & 0.189 & \underline{0.164} & \textbf{0.005} & 0.015 & \underline{0.008} & 0.013 \\
Turn$\to$Walk [L] & \textbf{0.643} & --- & --- & --- & \textbf{0.582} & --- & --- & --- & \textbf{0.096} & --- & --- & --- & \textbf{0.004} & --- & --- & --- \\
\midrule
\textit{Mean of $\pm$} & 0.007 & 0.016 & 0.025 & 0.019 & 0.077 & 0.065 & 0.139 & 0.090 & 0.011 & 0.017 & 0.029 & 0.018 & 0.003 & 0.003 & 0.003 & 0.002 \\
\textit{Std of $\pm$} & 0.003 & 0.011 & 0.012 & 0.011 & 0.044 & 0.044 & 0.053 & 0.039 & 0.008 & 0.022 & 0.018 & 0.014 & 0.003 & 0.002 & 0.002 & 0.002 \\
\bottomrule
\end{tabular}%
}
\end{table}

\subsection{Alignment, consistency, and warping}\label{sec:alignment}

Each family in Table~\ref{tab:main} is represented by a single variant, the one carrying the most transition information $I(T;Y)$ (Table~\ref{tab:information-loss}) and that variant is held fixed down its column. (Appendix~\ref{app:coverage}). Rows are the 11 transitions that at least two families recover and transitions found by only one family appear in Appendix~\ref{app:coverage}.

The DM family exhibits the strongest internal coherence on 9 of the 11 transitions. While alignment with annotated boundaries is more evenly distributed across methods, with the DM family achieving the best DTW scores on only 5 transitions, the warping decomposition reveals a clearer distinction. Specifically, the DM family achieves the lowest non-diagonal fraction on 9 of 11 transitions (Table~\ref{tab:main}) and the highest synchronization ratio on 9 of 11 (Table~\ref{tab:warping-dtw-sync}), whereas KPMS-KP does not lead on either metric. On each of the 4 rows where KPMS-KP has the lowest DTW its warping path is less diagonal than ours, so it reaches that DTW by stretching a primitive onto the annotated window, whereas the DM primitives follow the window at their own timescale. Two of the 11 rows are populated by DMC-Int alone among these columns; on the other 9 the DM family still leads on 7 for coherence, non-diagonal fraction and sync ratio alike.

On pure distribution metrics, DMC-Int trails KPMS-KP by $13\%$ on KID, which contrasts with the DTW result but in an informative way since the two metrics score different properties. DTW matches a primitive to its window under a monotone time correspondence, so it rewards recovering the order in which the body reconfigures, which is what a corridor or loop on the skeleton encodes and what the non-diagonal fraction confirms. KID is unchanged by permuting frames, so it rewards only covering the same region of pose space, where breadth helps: a KPMS-KP cluster holds a median of roughly $14{,}000$ frames against $4{,}000$ for DMP-Qpos, sampling the annotated window more fully while being markedly less coherent internally. The ordered structure our primitives capture is invisible to KID by construction, and we claim no distributional advantage. Coverage is likewise comparable, DMC-Int reaching 10 of the 16 discovered directed transitions against 11 for Mapper-Int and 9 for KPMS-KP (Appendix~\ref{app:coverage}).

\subsection{An information-theoretic view of coarse-graining}\label{sec:information}

\begin{figure}[!htbp]
    \centering
    \includegraphics[width=0.92\textwidth]{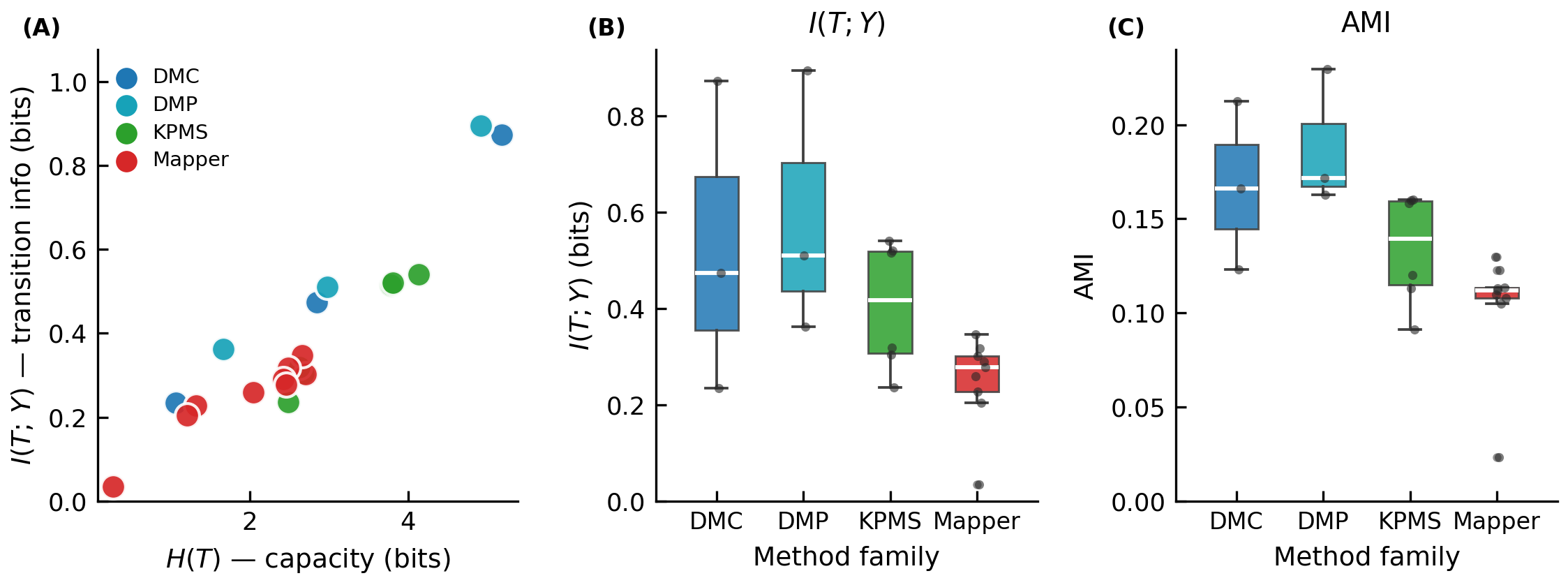}
    \caption{\textbf{How each coarse-graining preserves the transition signal.} \textbf{(A)}~Information Bottleneck plane: per-frame capacity $H(T)$ vs.\ transition information $I(T;Y)$ in bits, one dot per (method, seed); up-and-left dominates. \textbf{(B,C)}~Family-pooled $I(T;Y)$ and adjusted mutual information (AMI); the KPMS box is bimodal because KPMS-KP (high) and KPMS-Int (low) bracket the family. Per-variant numbers: Table~\ref{tab:information-loss}.}
    \label{fig:ib-panel}
\end{figure}

Beyond alignment, we ask how much of the transition signal each coarse-graining retains, computing the per-frame mutual information $I(T;Y)$ between the method's label $T_t$ and the directed-transition class $Y_t$ and placing each configuration on the Information Bottleneck plane~\citep{tishby2000information} (Figure~\ref{fig:ib-panel}A). DM cycle and path extraction in intention and qpos reaches the top of the plane, with the qpos DM-path carrying the most transition information of any method both before and after correcting label cardinality. DM methods reach this regime without an upstream state-label partition or a downstream pair-construction step. We read the gain through the same filter as in §\ref{sec:alignment}: a DM label is given only to frames on a persistent ridge, so it groups the cleaned part of the trajectory rather than every frame of the raw one. The within-DM ordering (intention $\approx$ qpos $>$ ekp) agrees with the alignment results. The information view shows that these features carry more transition signal than the best state-space pair-type construction while arising from a fundamentally different construction (per-variant numbers in Table~\ref{tab:information-loss}).

\section{Discussion and Limitations}\label{sec:discussion}

Several limitations remain. DM-G operates on individual ${\sim}500$-frame clips at 30~Hz, potentially missing transitions that span multiple clips. While pooling clips can recover these transitions (Appendix~\ref{app:stationary}), scalability is limited by the quadratic cost of sparse-Rips construction (Appendix~\ref{app:cost}). Finally, DM-G is not intended to compete with supervised transition classification in accuracy, but rather to provide a complementary, unsupervised, control-grounded representation of behavioral transitions (Appendix~\ref{app:supervised}).

\section{Conclusion}
Topological data analysis offers new tools for understanding behavioral dynamics. Existing pipelines coarse-grain the trajectory before ever asking how one behavior becomes the next, fixing a transition's describable resolution in advance. We instead reconstruct each trajectory's discrete-Morse skeleton in the control-grounded spaces of a deep-RL biomechanical emulator~\citep{mimicmjx2025}, a sparse subspace ``shaped by what is missing,'' and read two primitives off it: DM-paths through the corridors between dense behavioral regions, and DM-cycles around recurrences whose outgoing and returning routes differ. Because the skeleton keeps only the frames on persistent density ridges, it acts as a spatiotemporal filter on the raw feature space, much like recovering a road network from noisy car positions, so the primitives serve as an clean version of the trajectory. Clustering these primitives retains more transition information than baselines and matches the annotated boundaries with less warping and with more consistent clusters. Rather than replacing state-based methods, topology offers a complementary, transition-centric view of open-ended behavioral repertoires through their shape rather than the states into which they are partitioned.

\clearpage
\bibliography{bib}

\clearpage

\appendix
\section{Appendix}

\subsection{Discrete-Morse Based Graph Skeletonization Framework}\label{app:dm-details}

Discrete Morse theory was introduced by Forman~\citep{forman1998morse} as a combinatorial analog of Morse theory on cell complexes. Combined with persistent homology it yields a robust framework for extracting graph skeletons from a density field~\citep{delgado2015skeletonization,gyulassy2007topologically,robins2011theory,sousbie2011persistent,WangWangLi2015}: the skeleton is read off the ridges of the density terrain, and persistence retains only the topologically significant ridges while cancelling noise. \citet{DeyWangWang2018} formulate and simplify this with theoretical guarantees, and~\citet{MageeWang2022} extend it to high-dimensional point clouds. We extend it in turn to trajectories, and decompose the resulting skeletons to capture behavior transition patterns.

\begin{figure}[!htbp]
    \centering
    \includegraphics[width=\textwidth]{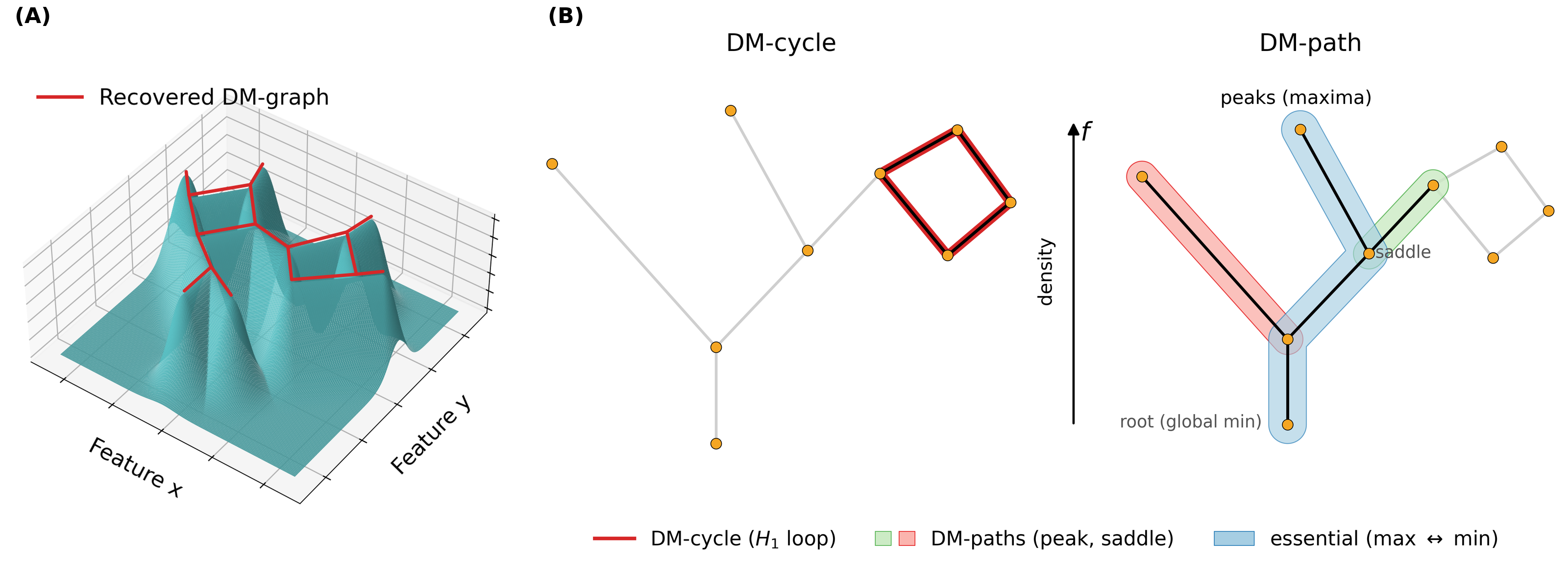}
    \caption{\textbf{Discrete-Morse skeletonization and the two primitives.} \textbf{(A)}~Skeletonization intuition: a near-uniform point cloud sampled around a hidden graph induces a density terrain, and the discrete-Morse reconstruction recovers the hidden graph (red) as the ridges of that terrain, the road-network intuition of §\ref{sec:setup}. \textbf{(B)}~The two DM primitives on one synthetic skeleton (a split tree plus one loop): the DM-cycle is the $H_1$ loop, and the DM-paths are the split-tree persistence pairs, each an up-fork leaf-to-saddle corridor (green, red), together with the essential global-max-to-global-min pair (blue). The skeleton is exactly the DM-cycle together with the DM-paths.}
    \label{fig:dmg-intuition}
\end{figure}

\S\ref{sec:setup} gives the road-network intuition; we make the underlying objects precise here. Let $f:\mathbb{R}^d\to\mathbb{R}$ be the density field, described for $d{=}2$ for concreteness, whose graph $\{(x,f(x))\}$ is the terrain of Figure~\ref{fig:dmg-intuition}A. We take the gradient to be the steepest \emph{descending} direction, the reverse of the usual convention but the one that makes flow lines easier to describe. \emph{Critical points} are those where it vanishes: maxima (peaks), minima (valley bottoms) and saddles. A drop of water released on the terrain traces a flow line downhill to a critical point; extending such lines uphill as well gives the \emph{integral lines} of $f$, which begin and end at critical points. The objects of interest are the \emph{1-stable manifolds}, the integral lines joining maxima to saddles: they connect density peaks through the saddles separating low-density basins, and their union is the graph we take as the hidden skeleton.

Being differential objects, 1-stable manifolds are hard to estimate reliably from discrete samples, the more so in high dimension. \citet{DeyWangWang2018} therefore work with the discrete Morse analog over simplicial complexes, where a simple combinatorial cancellation reduces the topological complexity of a discrete Morse function and so simplifies its 1-stable manifolds; persistent homology~\citep{edelsbrunner2010computational,dey2022computational} guides that cancellation so only topologically important features survive. See~\citet{DeyWangWang2018} for the algorithm and its guarantees.

\subsection{Feature Space Construction Details}\label{app:feature-spaces}

The three feature spaces (ekp, qpos, intention) are defined in §\ref{sec:setup}. The egocentric rotation removes the heading angle $\theta = \mathrm{atan2}(2(wz{+}xy),\, 1{-}2(y^2{+}z^2))$ and drops the root keypoint, leaving 20 keypoints $\times$ 3 = 60 dims. For qpos, the 7-dim root we strip is the global $(x, y, z)$ position plus the orientation quaternion. For ekp and qpos we additionally apply PCA to the leading 10 components to absorb residual allocentric variance into the discarded tail before running DM-G; intention's drift-invariance by construction makes this step unnecessary.

\begin{figure}[!htbp]
    \centering
    \includegraphics[width=\textwidth]{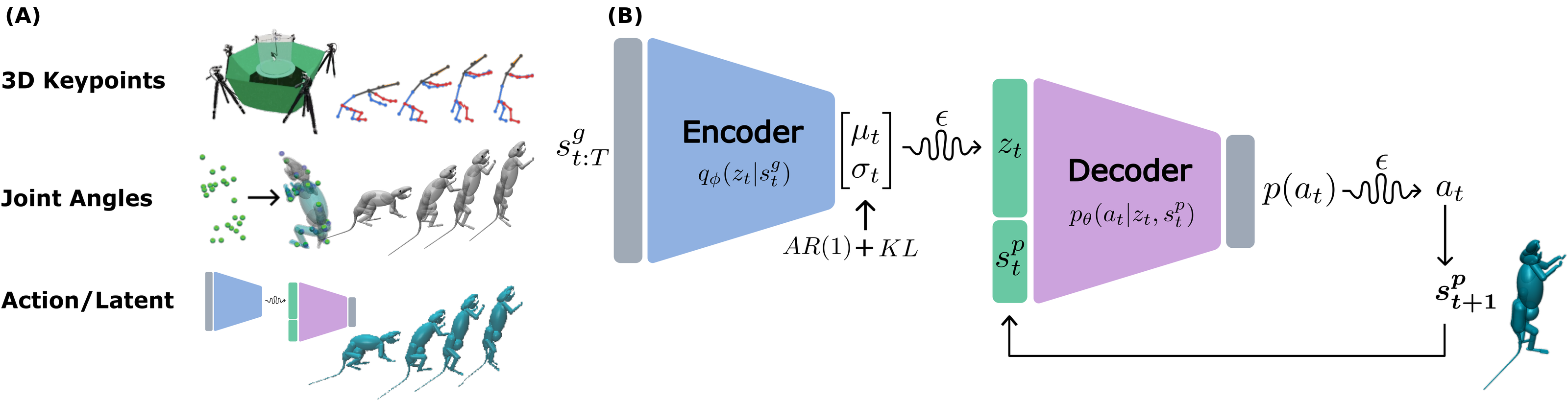}
    \caption{\textbf{The MIMIC-MJX pipeline that produces the trajectories analyzed here}~\citep{mimicmjx2025}. \textbf{(A)}~3D markerless motion capture is fit by inverse kinematics to a biomechanical model, and a deep-RL policy is trained to imitate the resulting joint trajectories. \textbf{(B)}~The encoder maps a short reference trajectory $s^{\text{ref}}_{t:t+5}$ into a low-dimensional latent intention $i_t$ under Gaussian and AR(1) regularization, and the decoder combines $i_t$ with the proprioceptive state $s_t$ to produce a biomechanically realistic action $a_t$.}
    \label{fig:mimic-pipeline}
\end{figure}

\subsection{Auto-Labeling Pipeline}\label{app:labeling}

\begin{figure}[h]
\centering
\includegraphics[width=\textwidth]{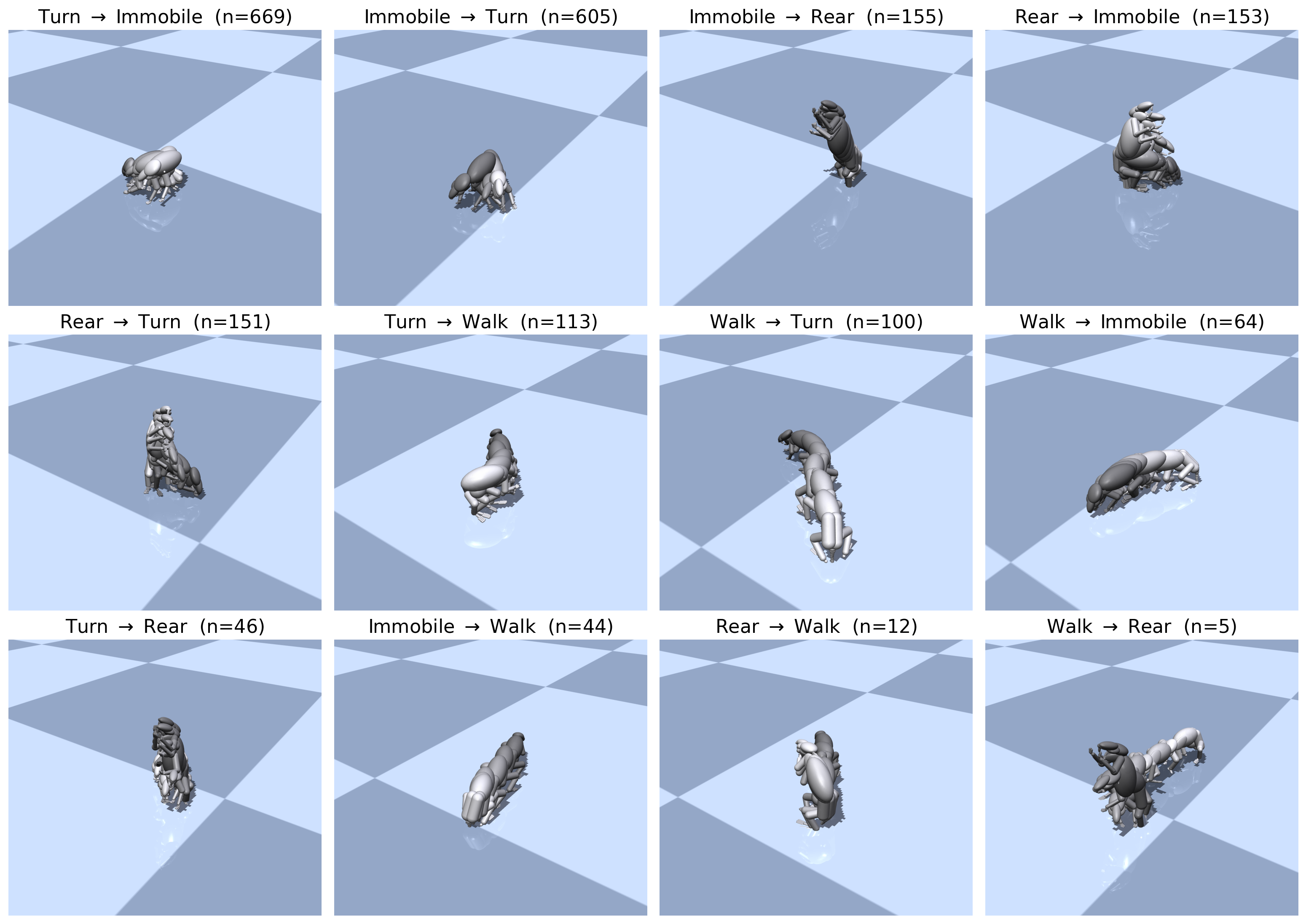}
\caption{\textbf{Rat behavior lexicon.} Each panel superimposes six poses of one representative ground-truth transition between two of the four macro-behaviors (Walk, Immobile, Turn, Rear), rendered in-scene from MJX rollouts with a shared camera, so the bodies occupy the positions the animal actually visited. Temporal progression is light-to-dark. Instances are selected by scoring each side of the labelled boundary against the kinematic signature of its own label (speed, yaw rate, torso height) and taking the weaker of the two, so both behaviors are legible rather than only the more energetic one. Panel titles give the directed transition type and its instance count $n$. Purely rotational transitions (\emph{Turn $\rightarrow$ Immobile}, \emph{Immobile $\rightarrow$ Turn}) remain the most compact, since a heading change at fixed position projects into self-occlusion from any single viewpoint.}
\label{fig:lexicon-rat}
\end{figure}

All ground-truth labels used in this paper are produced by an open-source heuristic auto-labeling pipeline\footnote{\url{https://github.com/KevinBian107/TopoMimic-annotation}} and then proofread by two annotators. Labels are used exclusively for evaluation, never during clustering. A representative instance of every directed transition is rendered in Figure~\ref{fig:lexicon-rat}.

\subsubsection{Rat Behavior Lexicon}

The pipeline labels every frame with a \emph{behavior} (one of four macro-categories) and a \emph{direction} (one of three turn modes), maintained as two parallel tracks per clip. The categories are defined by simple, interpretable kinematic heuristics applied over a 15-frame ($\sim$0.5\,s) sliding window:

\paragraph{Rear (R).} Triggered when the root height exceeds a fixed threshold or is rising rapidly relative to the moving baseline. Captures both supported and unsupported rears as a single category, since the MJX model lacks an environment surface and the two are visually indistinguishable in body coordinates.

\paragraph{Turn (T).} Triggered when planar rotational speed exceeds a threshold while planar translational speed stays below a translation cutoff. Captures stationary or low-velocity reorientations of the body axis.

\paragraph{Walk (W).} Triggered when planar translational speed exceeds the walk threshold. Captures all coordinated locomotion (walking and running are not separated in this version).

\paragraph{Immobile (I).} The default fallback when none of the above triggers. Captures stationary postures including grooming-like local limb activity that does not generate enough rotational or translational signal to fire one of the above triggers.

\paragraph{Priority and post-processing.} The four behaviors are assigned by first-match-wins priority \emph{Rear $>$ Turn $>$ Walk $>$ Immobile} (a frame can only carry one behavior). After per-frame assignment, two passes clean up flicker: (i) any segment shorter than 0.25\,s is absorbed into its longer neighbors, and (ii) any sandwiched ``A--B--A'' pattern with the inner B-segment shorter than 0.5\,s is bridged into a single ``A'' segment.

\paragraph{Direction track (Left / Right / Straight).} The direction label is computed independently from the behavior label by integrating the signed yaw rate of the body axis across the same window: positive integral above a turn-magnitude cutoff yields \emph{Left}, negative yields \emph{Right}, and below-cutoff integrals yield \emph{Straight}. The direction track is orthogonal to the behavior track, so any behavior can carry any direction (e.g., a rear can be straight, left, or right depending on the body's yaw during ascent).

\paragraph{Manual proofreading.} After auto-labeling, two annotators independently scan every clip and override frames where the heuristic clearly misfires (e.g., a slow drift labelled Walk that should be Immobile, or a rotational artifact during rearing). The proofreading is conservative: annotators only intervene when both reviewers agree the auto-label is wrong. The dataset used for evaluation throughout this paper reflects the post-proofread state of \texttt{behavior\_labels/rat\_labels.csv}.

\paragraph{Cluster-to-label assignment.} For evaluating a clustering method, each discovered cluster is assigned a transition label by majority vote over all of its members. Each member contributes the directed pair of the two behaviors that cover the most of its frames, ordered by first appearance (e.g., \emph{Rear $\rightarrow$ Walk}), together with the turn direction over the same frames: Left or Right if any frame carries that direction, the more frequent of the two winning and ties going to Left, and Straight otherwise. The most frequent (behavior pair, direction) tuple across the members becomes the cluster's ground-truth assignment, with ties broken alphabetically.

\subsection{Where the Primitives Appear}\label{app:primitive-checks}

The construction predicts that primitives appear exactly where behavior changes, and three checks bear this out: root displacement is larger for cycle-bearing clips than cycle-free ones, the longest DM-loop collapses on stationary clip pools while surviving on moving ones, and bootstrap-mean cycle counts are strictly higher for asymmetric A$\to$B$\to$A triplets than symmetric ones in every feature space (Appendices~\ref{app:stationary},~\ref{app:asymmetry}). Qualitatively, DM-G recovers both behaviors as density peaks (minima of $\rho$) and threads a corridor through the low-density band where their ground-truth frames meet, in every feature space (Figures~\ref{fig:appendix-transition-panel-tsne} and~\ref{fig:appendix-transition-filmstrips}, Appendix~\ref{app:transition-projections}). A clip may carry several persistent features or none, so we extract every cycle and path above a frame-count threshold and let the $W_2$ clustering absorb the multiplicity. The three subsections below give each check in full.

\subsection{Transition Projections Across Feature Spaces}\label{app:transition-projections}

\begin{figure}[h]
    \centering
    \includegraphics[width=\textwidth]{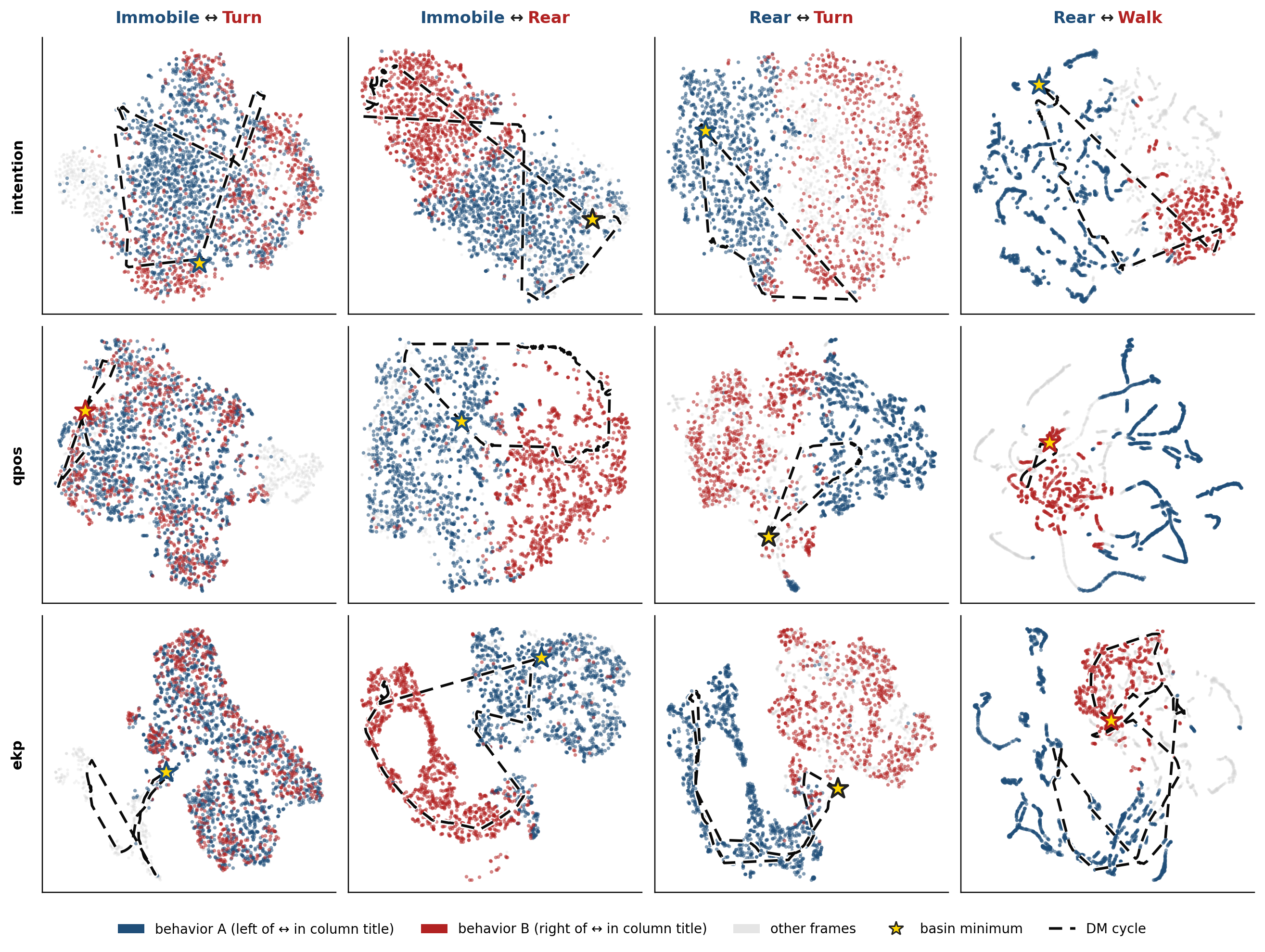}
    \caption{\textbf{DM-cycles $\times$ feature space.} Rows: intention / qpos / ekp; columns: the canonical transition pools. t-SNE of the Takens-embedded point cloud, with frames of the first behavior in blue, the second in red, and all other frames grey; gold stars mark the persistent density peaks (minima of $\rho$) DM-G identifies with the two behaviors, and the dashed black curve is the top DM-cycle of that pool, drawn closed. Intention and qpos recover clean loops for every pool; ekp, the most drift-laden space, recovers a loop in each but a visibly sparser one.}
    \label{fig:appendix-transition-panel-tsne}
\end{figure}

\begin{figure}[!htbp]
    \centering
    \includegraphics[width=\textwidth]{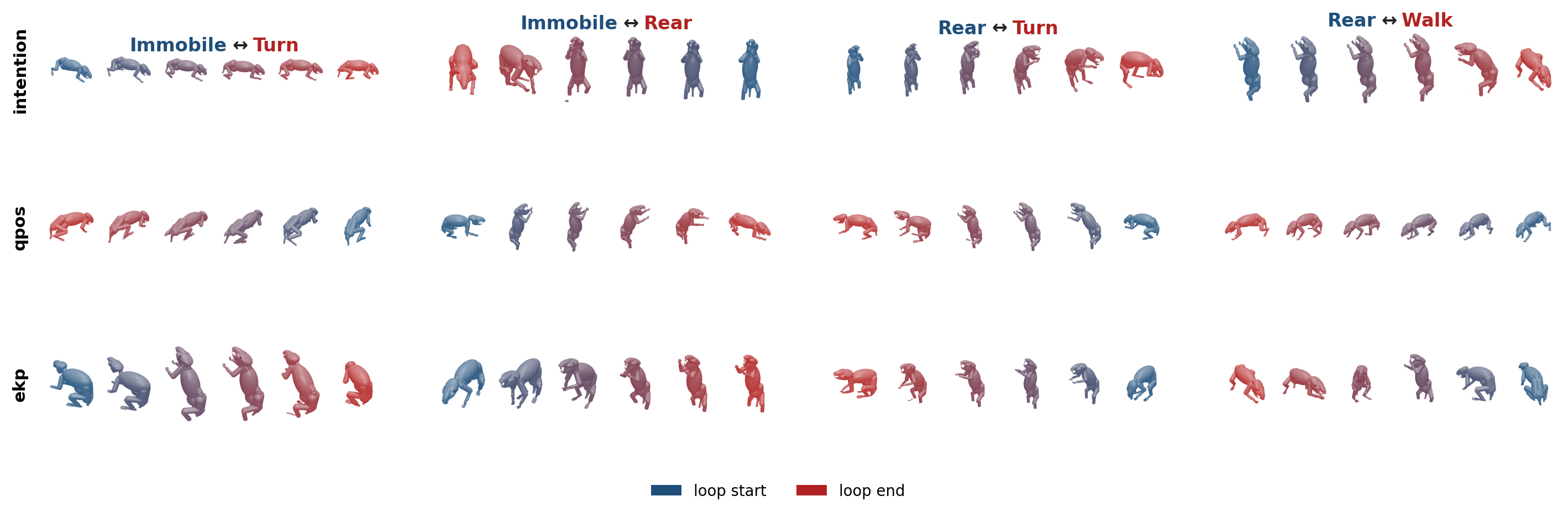}
    \caption{\textbf{DM-cycle filmstrips, all feature spaces.} Same layout as Figure~\ref{fig:appendix-transition-panel-tsne}: rows are feature spaces, columns the transition pools. Each strip shows the poses actually traversed around that pool's top DM-cycle, tinted light-to-dark by position along the loop, so the loop reads as a movement rather than as a curve in an embedding.}
    \label{fig:appendix-transition-filmstrips}
\end{figure}

Figures~\ref{fig:appendix-transition-panel-tsne} and~\ref{fig:appendix-transition-filmstrips} give the qualitative view of the DM-cycles referenced in §\ref{sec:results}, across all three feature spaces. For each of the canonical transition pools we take the Takens-embedded point cloud of the pooled clips, project it to two dimensions with t-SNE for visualisation only (the skeleton itself is built in the full embedding space, never on the projection), and overlay the extracted structure: frames labelled with the first behavior of the pair are drawn in blue, frames labelled with the second in red, all remaining frames in grey, gold stars mark the persistent density peaks (minima of $\rho$) that DM-G identifies with the two behaviors, and the dashed black curve is the top DM-cycle of that pool, drawn closed. Figure~\ref{fig:appendix-transition-filmstrips} then renders the poses actually traversed around each of those loops as a chronophotography filmstrip, tinted light-to-dark by position along the loop, so the loop can be read as a movement rather than as a curve in an embedding. Two things are visible. First, in every pool the two labelled behaviors occupy distinct regions and the recovered cycle passes through both, with the low-density corridor between them carrying the frames where the ground-truth transition is annotated: this is the qualitative signature behind the quantitative alignment results of §\ref{sec:alignment}. Second, the feature spaces differ in how cleanly they expose it. Intention and qpos recover clean, well-separated loops for every transition pool, whereas ekp, the most drift-laden space, still recovers a loop in each pool but a visibly sparser one, consistent with the drift argument of §\ref{subsec:primitives} and with ekp's weaker position in the alignment (Appendix~\ref{app:tables}) and transition-information (Appendix~\ref{app:information}) results.

\subsection{DM-G Identifies Stationary Clips}\label{app:stationary}

\begin{figure}[h]
    \centering
    \includegraphics[width=\textwidth]{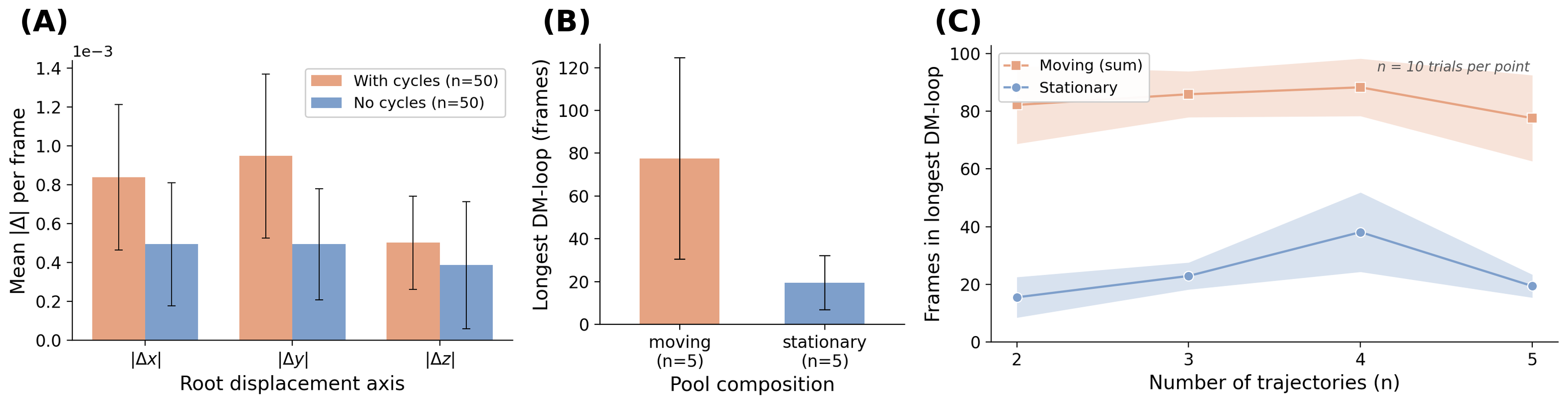}
    \caption{\textbf{DM-G identifies stationary clips.} \textbf{(A)}~Root displacement per frame ($|\Delta x|, |\Delta y|, |\Delta z|$) for clips with DM-cycles (orange) vs.\ without (blue), $n\!=\!50$ per group; cycle-bearing clips move more along every world axis. \textbf{(B)}~Longest DM-loop length on a $5$-clip moving (orange) vs.\ stationary (blue) pool: with no topological feature to encode, the skeleton collapses on the stationary pool. \textbf{(C)}~Same contrast swept across pool size $n\!\in\!\{2,3,4,5\}$, $10$ trials per point: pooled moving clips sustain a long loop across the full sweep while stationary-only pools stay flat near zero.}
    \label{fig:appendix-stationary}
\end{figure}

Of the 842 rat clips analyzed with DM-G in intention space, 602 produce at least one cycle (behaviorally rich) and 240 produce no cycles (stationary or topologically simple). Root displacement (mean $|\Delta|$ per frame, $n\!=\!50$ per group) is systematically larger along every world axis for cycle-bearing clips than for no-cycle clips (Figure~\ref{fig:appendix-stationary}A), so the cycles preferentially survive on behaviorally rich segments rather than emerging at random. The longest DM-loop length on a $5$-clip pool collapses to near-zero on a stationary-only composition while moving pools retain a long loop (Figure~\ref{fig:appendix-stationary}B), and the contrast scales smoothly across $n\!\in\!\{2,3,4,5\}$ pooled trajectories (Figure~\ref{fig:appendix-stationary}C), with a small bump at $n\!=\!4$ tracking the increased chance that a stationary pool happens to include a low-amplitude limb bout. The no-cycle population is dominated by Immobile frames, with the residual Turn / Walk / Rear mass corresponding to bouts that fall below DM-G's persistence threshold.

\subsection{Behavioral Asymmetry Across Feature Spaces}\label{app:asymmetry}

\begin{figure}[!htbp]
    \centering
    \includegraphics[width=\textwidth]{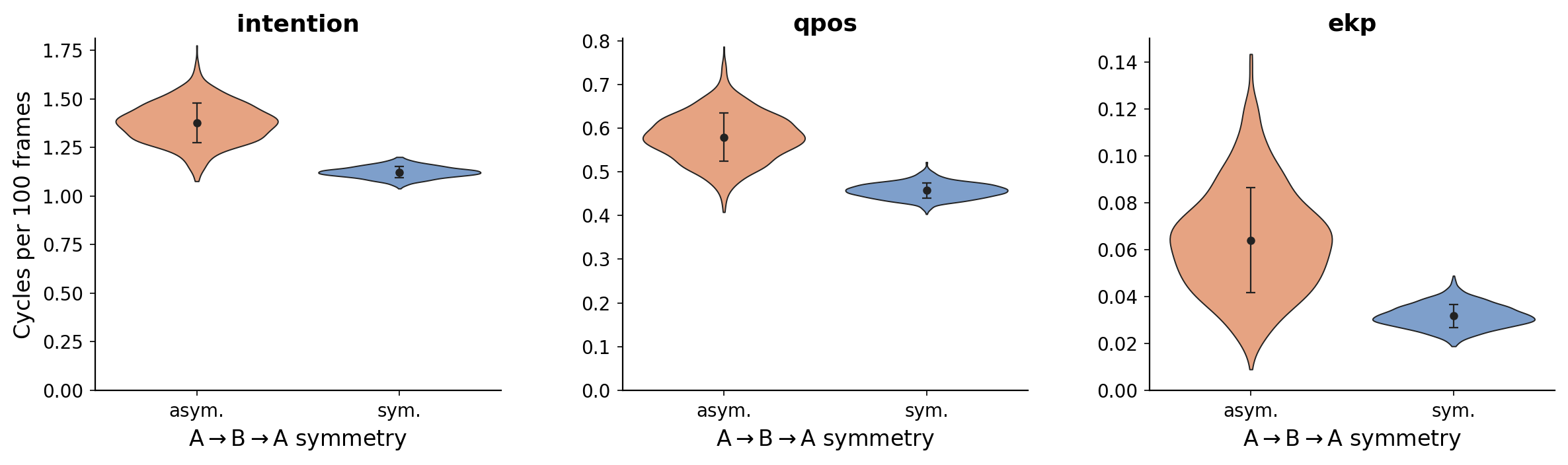}
    \caption{\textbf{Asymmetric A$\to$B$\to$A triplets carry strictly more cycles than symmetric ones across all feature spaces.} Violins are over bootstrap-mean ($n_{\mathrm{boot}}\!=\!1000$) cycle counts per $100$ frames. Orange = asymmetric (outgoing and returning routes differ in the dynamics), blue = symmetric. Mean and 95\% bootstrap percentile interval overlaid as black markers. Each panel uses its own y-axis because EKP cycle magnitudes are an order of magnitude smaller than intention / qpos.}
    \label{fig:appendix-asymmetry}
\end{figure}

A direct empirical consequence of the asymmetry intuition (§\ref{subsec:primitives}) is that an A$\to$B$\to$A triplet whose entry and exit arms occupy spatially distinct routes should carry more $H_1$ cycles than a symmetric triplet whose arms overlap. Bootstrapping the per-triplet cycle-count-per-$100$-frames ($n_{\mathrm{boot}}\!=\!1000$) across the rodent dataset matches this expectation in every feature space (Figure~\ref{fig:appendix-asymmetry}). Asymmetric triplets dominate symmetric ones at all three abstraction levels (intention, qpos, ekp), with the gap widening on the abstract intention space, consistent with the drift-invariance argument of §\ref{subsec:primitives}: intention does not have to allocate cycle budget to residual allocentric drift, so any cycle that does survive is more likely to be a behavioral asymmetry signal. We stress that these A$\to$B$\to$A triplets are only a convenient subset, chosen because a clear posture or position change makes the asymmetry easy to see; they are an experimental probe, not the full picture. Because the skeleton is built on the dynamics rather than on posture or position, other and more complex asymmetries can exist that this triplet test does not capture.

\subsection{Behavioral Coverage}\label{app:coverage}

Table~\ref{tab:coverage} reports coverage at the directed, direction-turn level, where Immobile$\to$Turn is counted separately from Turn$\to$Immobile and each is split by turn direction (left / right / straight). The label space nests: the four macro-behaviors give six unordered pairs, twelve directed transitions, and, tagged by turn direction, 36 classes, of which 25 have ground-truth instances in the evaluation split and 16 are discovered by at least one method; Table~\ref{tab:coverage} lists these 16, and Table~\ref{tab:main} reports the 11 of them for which more than one method family yields alignment metrics. Rear$\to$Walk [L], which appears in the per-variant alignment grids of Appendix~\ref{app:tables}, has no ground-truth instance in the evaluation split and is therefore not counted here. DM-WKC reaches transition coverage on par with the strongest existing models rather than exceeding them: DMC-Int discovers 10 of the 16 discovered directed transition classes, one fewer than Mapper-Int (11), with KPMS-KP at 9 and DMP-Int and DMP-Qpos at 8. Coverage breadth therefore does not by itself separate the methods; the DM advantage is in transition information (Table~\ref{tab:information-loss}) and lower warping (Table~\ref{tab:main}), not in the number of transitions discovered. The two families are not nested: four transitions, among them Walk$\to$Turn [L] and Walk$\to$Immobile [S], are recovered only by DM primitives (no baseline finds them at any tested setting), while Rear$\to$Immobile [R] is recovered only by Mapper, so each family exposes a few transitions the other misses. EKP-based DM methods sit at the narrow end of DM coverage (5 and 6 classes) because Cartesian keypoint coordinates retain the most residual allocentric drift after centering, washing out the density peaks for transitions that involve subtle postural changes (§\ref{subsec:primitives}).

\begin{table}[h]
\centering
\caption{\textbf{Behavioral coverage by method (directed, direction-turn).} Each cell indicates whether the method discovered ($\checkmark$) or missed ($\times$) at least one cluster majority-voted to the given \emph{directed} transition and turn direction (Immobile$\to$Turn is distinct from Turn$\to$Immobile). Direction tags: [L] left, [R] right, [S] straight. GT $n$ = ground-truth instances in the evaluation split. Only the 16 classes that at least one method discovers are listed; the evaluation split contains 25 classes with ground-truth instances.}
\label{tab:coverage}
\tiny
\setlength{\tabcolsep}{1.5pt}
\renewcommand{\arraystretch}{0.95}
\resizebox{\textwidth}{!}{%
\begin{tabular}{lrcccccccccccc}
\toprule
Transition & GT $n$ & \shortstack{DMC$^*$\\{\tiny (Int)}} & \shortstack{DMC$^*$\\{\tiny (Qpos)}} & \shortstack{DMC$^*$\\{\tiny (EKP)}} & \shortstack{DMP$^*$\\{\tiny (Int)}} & \shortstack{DMP$^*$\\{\tiny (Qpos)}} & \shortstack{DMP$^*$\\{\tiny (EKP)}} & \shortstack{Mapper\\{\tiny (Int)}} & \shortstack{Mapper\\{\tiny (Qpos)}} & \shortstack{Mapper\\{\tiny (EKP)}} & \shortstack{KPMS\\{\tiny (Int)}} & \shortstack{KPMS\\{\tiny (KP)}} & \shortstack{Full\\{\tiny (Int)}} \\
\midrule
Immobile$\rightarrow$Rear [S] & 31 & $\checkmark$ & $\checkmark$ & $\checkmark$ & $\times$ & $\checkmark$ & $\checkmark$ & $\checkmark$ & $\checkmark$ & $\checkmark$ & $\checkmark$ & $\checkmark$ & $\checkmark$ \\
Immobile$\rightarrow$Rear [L] & 4 & $\times$ & $\checkmark$ & $\checkmark$ & $\checkmark$ & $\checkmark$ & $\checkmark$ & $\checkmark$ & $\times$ & $\times$ & $\times$ & $\checkmark$ & $\checkmark$ \\
Immobile$\rightarrow$Rear [R] & 2 & $\times$ & $\times$ & $\checkmark$ & $\times$ & $\checkmark$ & $\checkmark$ & $\times$ & $\times$ & $\times$ & $\times$ & $\checkmark$ & $\checkmark$ \\
Immobile$\rightarrow$Turn [L] & 95 & $\checkmark$ & $\checkmark$ & $\times$ & $\checkmark$ & $\checkmark$ & $\times$ & $\checkmark$ & $\checkmark$ & $\checkmark$ & $\checkmark$ & $\checkmark$ & $\checkmark$ \\
Immobile$\rightarrow$Turn [R] & 45 & $\checkmark$ & $\checkmark$ & $\times$ & $\checkmark$ & $\checkmark$ & $\times$ & $\checkmark$ & $\checkmark$ & $\checkmark$ & $\times$ & $\checkmark$ & $\checkmark$ \\
Immobile$\rightarrow$Walk [L] & 1 & $\checkmark$ & $\times$ & $\times$ & $\times$ & $\times$ & $\times$ & $\checkmark$ & $\times$ & $\times$ & $\times$ & $\times$ & $\times$ \\
Rear$\rightarrow$Immobile [S] & 26 & $\checkmark$ & $\times$ & $\checkmark$ & $\checkmark$ & $\times$ & $\checkmark$ & $\checkmark$ & $\checkmark$ & $\checkmark$ & $\checkmark$ & $\checkmark$ & $\times$ \\
Rear$\rightarrow$Immobile [L] & 4 & $\times$ & $\checkmark$ & $\checkmark$ & $\times$ & $\checkmark$ & $\checkmark$ & $\checkmark$ & $\checkmark$ & $\checkmark$ & $\times$ & $\checkmark$ & $\checkmark$ \\
Rear$\rightarrow$Immobile [R] & 2 & $\times$ & $\times$ & $\times$ & $\times$ & $\times$ & $\times$ & $\checkmark$ & $\times$ & $\checkmark$ & $\times$ & $\times$ & $\times$ \\
Rear$\rightarrow$Turn [L] & 26 & $\checkmark$ & $\times$ & $\times$ & $\checkmark$ & $\checkmark$ & $\checkmark$ & $\checkmark$ & $\checkmark$ & $\times$ & $\checkmark$ & $\checkmark$ & $\times$ \\
Rear$\rightarrow$Turn [R] & 6 & $\checkmark$ & $\times$ & $\times$ & $\times$ & $\times$ & $\times$ & $\times$ & $\times$ & $\times$ & $\times$ & $\times$ & $\times$ \\
Turn$\rightarrow$Immobile [L] & 102 & $\checkmark$ & $\checkmark$ & $\times$ & $\checkmark$ & $\checkmark$ & $\times$ & $\checkmark$ & $\checkmark$ & $\times$ & $\times$ & $\checkmark$ & $\checkmark$ \\
Turn$\rightarrow$Rear [L] & 6 & $\checkmark$ & $\times$ & $\times$ & $\times$ & $\times$ & $\times$ & $\times$ & $\times$ & $\times$ & $\times$ & $\times$ & $\times$ \\
Turn$\rightarrow$Walk [L] & 22 & $\checkmark$ & $\times$ & $\times$ & $\times$ & $\times$ & $\times$ & $\checkmark$ & $\times$ & $\times$ & $\checkmark$ & $\times$ & $\times$ \\
Walk$\rightarrow$Immobile [S] & 11 & $\times$ & $\times$ & $\times$ & $\checkmark$ & $\times$ & $\times$ & $\times$ & $\times$ & $\times$ & $\times$ & $\times$ & $\times$ \\
Walk$\rightarrow$Turn [L] & 17 & $\times$ & $\times$ & $\times$ & $\checkmark$ & $\times$ & $\times$ & $\times$ & $\times$ & $\times$ & $\times$ & $\times$ & $\times$ \\
\midrule
\textit{Total discovered} & & 10 & 6 & 5 & 8 & 8 & 6 & 11 & 7 & 6 & 5 & 9 & 7 \\
\bottomrule
\end{tabular}%
}
\vspace{2pt}
\\\scriptsize{$^*$Ours.}
\end{table}

\subsection{Controls}\label{app:controls}

Three controls rule out simpler explanations. First, none of 110 variants spanning KPMS stickiness and state count or Mapper lens and interval count improves on the tuned baselines (Appendix~\ref{app:scale-sweep}). Second, the full-clip control, which omits decomposition, performs worse than both DM variants in alignment (mean DTW $0.888$) and coherence ($0.797$), showing that decomposition rather than Wasserstein clustering recovers the transitions. Third, DM-cycle DTW varies by at most $0.03$ across $k$ in every feature space, while persistence pruning trades transition information for cluster quality without a uniformly better operating point; we therefore retain all paths (Appendices~\ref{app:k-sweep} and~\ref{app:path-persistence}). The primitives are also complementary: DM-paths resolve kinematically similar modes, whereas DM-cycles capture transitions involving larger postural changes.

\subsection{Baseline Scale Sensitivity}\label{app:scale-sweep}

State-space and topological baselines expose timescale-related knobs that directly control how long a typical behavioral segment lasts: KPMS's stickiness $\kappa$ and number of states $N_{\text{states}}$ jointly set the expected syllable dwell time, and Mapper's number of cover intervals $n_{\text{intervals}}$ sets how finely the lens partitions the manifold. To characterize how baseline coverage depends on these knobs rather than assuming a single setting, we sweep these hyperparameters and ask, for each variant, how many of the 16 directed, direction-turn transition classes of Table~\ref{tab:coverage} it discovers, scored over the same rows so the two tables share a denominator.

For KPMS we sweep $\kappa \in \{10^{3}, 10^{4}, 10^{5}, 10^{6}, 10^{7}\}$ against $N_{\text{states}} \in \{5,10,15,20,30\}$ on \emph{both} feature streams, 50 fits in total. Every fit is scored for discovery; we deliberately apply no model-selection criterion first, since the question here is whether these knobs limit baseline coverage at all, which is a question about the union over settings rather than about any single preferred setting. Both streams use the heading treatment of Appendix~\ref{app:baseline-hp}. For Mapper we pair four lenses (Fiedler, eccentricity, kNN-density, and PCA-1D) with three feature spaces (intention, qpos, ekp) and sweep $n_{\text{intervals}} \in \{5, 10, 15, 20, 25\}$, yielding 60 Mapper variants. The remaining Mapper settings are held fixed across the sweep at overlap fraction $0.5$, DBSCAN $\texttt{min\_samples}{=}5$, and the same global neighbourhood-radius rule as Appendix~\ref{app:mapper-hp}; nodes covering fewer than 500 frames in total are pruned before discovery is scored, which the benchmark configuration of Appendix~\ref{app:mapper-hp} does not do. The sweep therefore characterizes sensitivity to the cover and the lens, and not to the inner-clustering scale. A transition $A{\leftrightarrow}B$ counts as \emph{discovered} if the variant produces at least one cluster whose frame-level ground-truth purity for that transition exceeds a fixed threshold and whose member-clip count clears a minimum support size. Table~\ref{tab:scale-sweep} reports the count for every one of the 110 variants.

Cells count the 16 directed, direction-turn classes of Table~\ref{tab:coverage}, with primitives built and labelled exactly as for the benchmark baselines. Discovery stays low across the whole grid: no intention fit recovers more than 2 of the 16 classes, the keypoint stream recovers none at $\kappa{=}10^{3}$ for any state count, and no fit with $N_{\text{states}}{=}5$ recovers more than one class on either stream. The best single KPMS variant reaches 4 of the 16 classes and the best single Mapper variant 6, the latter on ekp with the Fiedler lens, while the union over all 50 KPMS variants reaches 10 and the union over all 60 Mapper variants 9. Decisively, no swept setting improves on the tuned baseline configurations of Appendix~\ref{app:baseline-hp}, which reach 9 of 16 for KPMS and 11 for Mapper in Table~\ref{tab:coverage}: the sweep holds every remaining setting at a common default rather than at those per-stream tuned values, and varying the timescale knobs around it never recovers the difference. The configurations reported in Table~\ref{tab:coverage} therefore remain the strongest baselines we obtained, and stickiness and state count are not hiding a better one. Over the settings explored the sweep therefore shows that the baselines' coverage is not primarily an artifact of these knobs; it also holds Mapper's DBSCAN neighbourhood radius fixed, so it bounds coverage over the cover and lens rather than over every hyperparameter.

\begin{table}[h]
\centering
\caption{\textbf{Baseline scale-hyperparameter sweep.} Each cell is the number of the 16 directed, direction-turn transition classes of Table~\ref{tab:coverage} that the variant discovers, over the same rows and the same evaluation split, so the two tables share a denominator. \textbf{(a)} Keypoint-MoSeq over stickiness $\kappa$ and $N_{\text{states}}$ on both feature streams (50 fits). \textbf{(b)} Mapper over lens and $n_{\text{intervals}}$ on all three feature spaces (60 variants). Primitives are built exactly as for the benchmark baselines, from sustained syllable or node pairs, and each cluster is majority-voted to a class by the same labeller used for Table~\ref{tab:coverage}. Every fit is scored with no model-selection criterion applied first, so the union line is a true oracle union over all 110 variants. Remaining settings are held at a common default rather than at the per-stream tuned values of Appendix~\ref{app:baseline-hp}, and no swept setting improves on the tuned configurations reported in Table~\ref{tab:coverage}.}
\label{tab:scale-sweep}

\begin{minipage}[t]{0.46\textwidth}
\centering
\textit{(a) Keypoint-MoSeq.}\\[2pt]
{\scriptsize
\setlength{\tabcolsep}{3pt}
\renewcommand{\arraystretch}{0.95}
\begin{tabular}{llccccc}
\toprule
& & \multicolumn{5}{c}{$N_{\text{states}}$} \\
\cmidrule(lr){3-7}
Feature space & $\kappa$ & 5 & 10 & 15 & 20 & 30 \\
\midrule
\multicolumn{7}{l}{\textit{Intention}} \\
\quad & $10^{3}$ & 0 & 2 & 2 & 2 & 2 \\
\quad & $10^{4}$ & 1 & 2 & 1 & 1 & 1 \\
\quad & $10^{5}$ & 1 & 1 & 0 & 1 & 1 \\
\quad & $10^{6}$ & 1 & 0 & 0 & 2 & 0 \\
\quad & $10^{7}$ & 1 & 1 & 0 & 1 & 0 \\
\multicolumn{7}{l}{\textit{Keypoint}} \\
\quad & $10^{3}$ & 0 & 0 & 0 & 0 & 0 \\
\quad & $10^{4}$ & 1 & 3 & 1 & 1 & 2 \\
\quad & $10^{5}$ & 0 & 2 & 3 & 0 & 4 \\
\quad & $10^{6}$ & 0 & 3 & 3 & 2 & 2 \\
\quad & $10^{7}$ & 1 & 2 & 3 & 3 & 2 \\
\bottomrule
\end{tabular}
}
\end{minipage}\hfill
\begin{minipage}[t]{0.52\textwidth}
\centering
\textit{(b) Mapper.}\\[2pt]
{\scriptsize
\setlength{\tabcolsep}{3pt}
\renewcommand{\arraystretch}{0.95}
\begin{tabular}{llccccc}
\toprule
& & \multicolumn{5}{c}{$n_{\text{intervals}}$} \\
\cmidrule(lr){3-7}
Feature space & Lens & 5 & 10 & 15 & 20 & 25 \\
\midrule
\multicolumn{7}{l}{\textit{Intention}} \\
\quad & Fiedler & 0 & 2 & 0 & 0 & 1 \\
\quad & kNN-density & 0 & 0 & 0 & 0 & 0 \\
\quad & PCA-1D & 1 & 0 & 2 & 3 & 1 \\
\quad & Eccentricity & 0 & 1 & 0 & 1 & 0 \\
\multicolumn{7}{l}{\textit{Qpos}} \\
\quad & Fiedler & 3 & 4 & 2 & 1 & 2 \\
\quad & kNN-density & 0 & 3 & 2 & 1 & 0 \\
\quad & PCA-1D & 0 & 0 & 2 & 4 & 3 \\
\quad & Eccentricity & 0 & 3 & 1 & 0 & 0 \\
\multicolumn{7}{l}{\textit{EKP}} \\
\quad & Fiedler & 2 & 4 & 3 & 6 & 3 \\
\quad & kNN-density & 0 & 0 & 1 & 2 & 0 \\
\quad & PCA-1D & 1 & 2 & 1 & 1 & 2 \\
\quad & Eccentricity & 0 & 0 & 0 & 1 & 0 \\
\bottomrule
\end{tabular}
}
\end{minipage}

\vspace{4pt}

\scriptsize{Oracle union over all 50 KPMS variants: 10 of 16. Over all 60 Mapper variants: 9 of 16. Combined baseline union: 11 of 16.}
\end{table}

\subsection{Trajectory-to-Skeleton Reduction}\label{app:reduction}

Table~\ref{tab:reduction} reports the size of a trajectory at each stage of the pipeline, per feature space: the original clip length, the DM-skeleton it is summarized to, and the lengths of the DM-cycle and DM-path primitives the skeleton decomposes into. A raw clip of $n \approx 500$ frames yields a skeleton of about $263$ (intention), $344$ (qpos), and $260$ (ekp) vertices, and Step 2 cuts that skeleton into primitives that are short sub-trajectories: DM-cycles average $75$, $113$, and $191$ frames and DM-paths $62$, $113$, and $105$ frames across the three spaces, so each captures a few seconds of behavior rather than the whole clip. Skeleton sizes are averaged over the clips that produce a non-empty skeleton, the primitive lengths over all extracted primitives.

Before clustering, every primitive is resampled to a fixed length of $L = 100$ frames (Step 3, §\ref{sec:cluster}), down-sampling the longer ones and interpolating the shorter ones up, so that all are compared as equal-size empirical distributions regardless of their native span.

\begin{table}[!htbp]
    \centering
    \small
    \caption{\textbf{Trajectory-to-skeleton reduction by feature space.} Mean over rat clips of the original trajectory length (Frames), the DM-skeleton vertex count (Skeleton), and the per-primitive lengths of the DM-cycles and DM-paths. Frames and both primitive lengths are in frames; Skeleton is in graph vertices. Skeleton is averaged over clips that yield a non-empty skeleton, the primitive lengths over all extracted primitives. Each primitive is resampled to $L=100$ frames before clustering (§\ref{sec:cluster}).}
    \label{tab:reduction}
\begin{tabular}{lcccc}
\toprule
Feature space & Frames & Skeleton & DM-cycle & DM-path \\
\midrule
Intention & 499 & 263 & 75 & 62 \\
Qpos & 500 & 344 & 113 & 113 \\
EKP & 500 & 260 & 191 & 105 \\
\bottomrule
\end{tabular}
\end{table}

\subsection{Wasserstein Clustering of Primitives}\label{app:wkc}

§\ref{sec:cluster} states the DM-WKC construction; here we record what it takes to reproduce it. Primitives run from roughly $60$ to $190$ frames (Appendix~\ref{app:reduction}), so each is resampled to $L = 100$ before any distance is computed, keeping evenly spaced frames when a primitive is longer than $L$ and interpolating frame indices when it is shorter, with primitives of fewer than three frames discarded; clip and frame indices travel with the resampled primitive, so a cluster can be mapped back to the frames it was built from, which is what Appendix~\ref{app:cluster-examples} renders. Because both measures then carry $L$ atoms of equal mass $1/L$, an optimal plan may be taken to be a permutation and the transport cost is a minimum-cost perfect matching between the two point sets, which we solve exactly by network simplex~\citep{flamary2021pot} rather than by an entropic approximation, under a squared Euclidean ground cost, so the entries of $\mathbf{D}$ are squared $2$-Wasserstein distances. One consequence is that $\mu_j$ is unordered: the transport is computed on the per-frame feature vectors $p_t \in \mathcal{Z}$ that a primitive visits rather than on their Takens embeddings, so the distance does not see the order in which a primitive is traversed, and temporal structure enters only through which frames the skeleton assigns to the primitive. For qpos and ekp these are the root-stripped and egocentric features of §\ref{sec:background}, without the PCA applied before skeletonization. The matrix is then rescaled entrywise by $\sqrt{r_i c_j}$, with $r_i$ and $c_j$ the median off-diagonal distances of row $i$ and column $j$, putting every primitive's distances on a comparable scale, before metric MDS on precomputed dissimilarities and $K$-means with $K = 25$ (Appendix~\ref{app:dmg-hp}). We cache $\mathbf{D}$, so the embedding dimension can be swept (Appendix~\ref{app:mds-sweep}) without recomputing any transport; those $O(N^2)$ solves dominate the cost of this stage (Appendix~\ref{app:cost}).

\subsection{Discrete-Morse $k$-Neighbor Sensitivity}\label{app:k-sweep}

The skeleton is built from a $k$-nearest-neighbor density estimate (§\ref{subsec:DMgraph}), so $k$ sets how strongly the density is smoothed before simplification. We sweep $k \in \{3, 5, 8\}$ ($k{=}3$ shipped) on all three feature spaces, holding every other setting at its production value and rerunning the full pipeline per $k$ so the numbers stay comparable to the main results. Table~\ref{tab:k-sweep} reports the label-free skeleton structure (DM-cycles per clip and giant-connected-component fraction) and the DM-cycle alignment (DTW, KID, coverage).

The headline metric is essentially flat: DTW varies by at most $0.03$ across $k$ in every space. The only clear directional effect favors small $k$ where cycles carry the most signal. In intention, raising $k$ from $3$ to $8$ erodes the cycle structure (DM-cycles per clip: $2.20\!\to\!0.83$, giant-component fraction: $0.51\!\to\!0.25$) and worsens the distributional metric (KID: $0.009\!\to\!0.017$) with no offsetting gain, and the same KID degradation appears in ekp ($0.010\!\to\!0.025$); qpos drifts mildly the other way in connectivity while its alignment stays flat. Larger $k$ buys only marginally more coverage (intention: $6.0\!\to\!7.3$) at equal or worse alignment. We therefore ship $k{=}3$, which recovers the richest topology and is best in intention on both structure and alignment. Since $k$ and $\delta$ jointly shape the skeleton, this characterizes sensitivity at each space's shipped $\delta$ rather than over the full $(k, \delta)$ grid.

\begin{table}[!htbp]
    \centering
    \caption{\textbf{Discrete-Morse $k$-neighbor sensitivity on rat transitions.} Per feature space and kNN parameter $k$ (shipped value $k{=}3$ in bold), the label-free skeleton structure (mean number of DM-cycle primitives per clip, counted after temporal splitting, and the fraction of all embedded points in the largest connected component of the skeleton) and the DM-cycle ground-truth alignment (DTW, KID, and number of discovered transitions). DTW is flat across $k$ in every space; larger $k$ erodes the intention cycle structure and its KID with no alignment gain.}
    \label{tab:k-sweep}
\begin{tabular}{llcccccc}
\toprule
Space & $k$ & cycles/clip & giant-CC & DTW & KID & cov \\
\midrule
Intention & \textbf{3} & 2.20 & 0.51 & 0.80 & 0.009 & 6.0 \\
 & 5 & 1.59 & 0.40 & 0.83 & 0.014 & 6.3 \\
 & 8 & 0.83 & 0.25 & 0.82 & 0.017 & 7.3 \\
\midrule
Qpos & \textbf{3} & 0.83 & 0.41 & 0.85 & 0.008 & 5.0 \\
 & 5 & 0.75 & 0.44 & 0.83 & 0.009 & 5.3 \\
 & 8 & 0.67 & 0.46 & 0.83 & 0.009 & 5.7 \\
\midrule
EKP & \textbf{3} & 0.19 & 0.13 & 0.94 & 0.010 & 4.0 \\
 & 5 & 0.16 & 0.12 & 0.93 & 0.015 & 4.0 \\
 & 8 & 0.14 & 0.10 & 0.93 & 0.025 & 4.7 \\
\bottomrule
\end{tabular}
\end{table}

\subsection{Path-Selection Persistence Sensitivity}\label{app:path-persistence}

The DM-path decomposition (§\ref{subsec:primitives}) keeps every peak-to-saddle path plus each tree's global-max-to-global-min path in density (§\ref{subsec:primitives}); the shipped $\delta_{\mathrm{path}}=0$ keeps all of them. A threshold $\delta_{\mathrm{path}}>0$ discards the low-persistence paths, the minor density peaks that merge into a higher one through a saddle of nearly the same density, leaving only deep, well-separated corridors. We sweep $\delta_{\mathrm{path}}$ as a multiple of each space's construction threshold $\delta$; because this only re-filters branches of the already-built skeleton, it reuses the cached skeleton with no DM-G rerun. We score both the transition information of the path labeling ($I(T;Y)$ and AMI; Table~\ref{tab:path-persistence-ami}) and the downstream DMP clustering alignment (Table~\ref{tab:path-persistence-down}).

The two views disagree on the optimum. Transition information falls monotonically as $\delta_{\mathrm{path}}$ rises, in every space and even for the cardinality-corrected AMI, so the discarded paths carry signal rather than noise: a low-persistence path is a short, localized mode-to-boundary corridor and is highly transition-specific, whereas the high-persistence backbone spans the whole trajectory and is less so, so keeping all paths maximizes raw information. Downstream, however, a mild threshold helps the drift-clean spaces: at $\delta_{\mathrm{path}}{=}\delta$, intention and qpos drop KID by $46$ and $17$ percent with DTW unchanged and coverage equal or higher (removing short noisy branches tightens the clusters), while ekp gains little (KID $0.014\!\to\!0.013$, DTW $0.916\!\to\!0.924$). The shipped $\delta_{\mathrm{path}}=0$ thus sits at the information-optimal end and is safe everywhere, with $\delta_{\mathrm{path}}\approx\delta$ a defensible alternative when downstream cluster quality in intention and qpos is the priority.

\begin{table}[!htbp]
    \centering
    \caption{\textbf{Path-selection persistence: transition information.} Per feature space and path-persistence multiplier $\delta_{\mathrm{path}}/\delta$ (shipped $0$ in bold), the number of DM-paths and the transition information of the per-frame DMP coarse-graining: mutual information $I(T;Y)$, cardinality-corrected AMI, and label entropy $H(T)$. All three decrease monotonically as the threshold rises.}
    \label{tab:path-persistence-ami}
\begin{tabular}{llcccc}
\toprule
Space & $\delta_{\mathrm{path}}/\delta$ & paths & $I(T;Y)$ & AMI & $H(T)$ \\
\midrule
Intention & \textbf{0} & 1389 & 0.511 & 0.172 & 2.98 \\
 & 0.5 & 1333 & 0.495 & 0.169 & 2.90 \\
 & 1 & 1201 & 0.462 & 0.163 & 2.70 \\
 & 2 & 881 & 0.368 & 0.146 & 2.13 \\
 & 4 & 825 & 0.346 & 0.141 & 2.00 \\
\midrule
Qpos & \textbf{0} & 1140 & 0.896 & 0.230 & 4.92 \\
 & 0.5 & 942 & 0.827 & 0.223 & 4.55 \\
 & 1 & 796 & 0.757 & 0.219 & 4.10 \\
 & 2 & 562 & 0.552 & 0.190 & 3.02 \\
 & 4 & 553 & 0.528 & 0.183 & 2.97 \\
\midrule
EKP & \textbf{0} & 413 & 0.364 & 0.163 & 1.67 \\
 & 0.5 & 337 & 0.322 & 0.151 & 1.49 \\
 & 1 & 290 & 0.281 & 0.138 & 1.32 \\
 & 2 & 253 & 0.249 & 0.126 & 1.20 \\
 & 4 & 245 & 0.242 & 0.123 & 1.18 \\
\bottomrule
\end{tabular}
\end{table}

\begin{table}[!htbp]
    \centering
    \caption{\textbf{Path-selection persistence: downstream alignment.} DMP clustering alignment (DTW, KID, and discovered-transition coverage) at three path-persistence multipliers ($0$ = shipped, in bold). A mild threshold ($\delta_{\mathrm{path}}{=}\delta$) lowers KID and raises coverage in intention and qpos with DTW unchanged, while ekp is roughly unchanged.}
    \label{tab:path-persistence-down}
\begin{tabular}{llccc}
\toprule
Space & $\delta_{\mathrm{path}}/\delta$ & DTW & KID & cov \\
\midrule
Intention & \textbf{0} & 0.758 & 0.013 & 2.7 \\
 & 1 & 0.755 & 0.007 & 3.0 \\
 & 4 & 0.801 & 0.010 & 3.0 \\
\midrule
Qpos & \textbf{0} & 0.840 & 0.013 & 4.3 \\
 & 1 & 0.842 & 0.011 & 6.3 \\
 & 4 & 0.838 & 0.010 & 6.3 \\
\midrule
EKP & \textbf{0} & 0.916 & 0.014 & 5.7 \\
 & 1 & 0.924 & 0.013 & 6.0 \\
 & 4 & 0.915 & 0.015 & 6.3 \\
\bottomrule
\end{tabular}
\end{table}

\subsection{Information-Theoretic Analysis of Coarse-Graining}
\label{app:information}

We evaluate each unsupervised behavioral coarse-graining by measuring how much information its per-frame output retains about an external relevance variable. The relevance variable is the ground-truth directed-transition class, and we report mutual information together with normalized and cardinality-corrected variants. All variables are defined on the labeled subset of the rodent dataset that contains at least one valid transitions (578/842 clips). For frame $t$ in clip $c$, $Y_t \in \{1,\ldots,12\}$ denotes the directed transition class (produced by the auto-labeling pipeline in Appendix~\ref{app:labeling}). This variable is method-agnostic and has marginal entropy $H(Y)=2.67$ bits, which upper-bounds any achievable $I(T;Y)$. The coarse-graining label $T_t$ is the method-specific integer primitive assigned to frame $t$, with $T_t=-1$ reserved for unassigned frames. The construction of $T_t$ differs by method:

\begin{itemize}
    \item \textbf{DM-cycles (DMC):} $T_t$ is the global index, across all clips, of the discrete-Morse cycle containing frame $t$, or $-1$ if no cycle contains it.
    \item \textbf{DM-paths (DMP):} $T_t$ is defined analogously using descending-path primitives.
    \item \textbf{KPMS:} $T_t$ is the retained ordered pair type of source and target syllables for the sustained-transition window containing frame $t$.
    \item \textbf{Mapper:} $T_t$ is the analogous ordered pair type, formed exactly as for KPMS but over Mapper graph-node labels instead of syllables (Appendix~\ref{app:mapper-hp}).
\end{itemize}

This definition preserves each method's native transition unit. DM methods produce instance-level topological objects, so DMC and DMP have hundreds of labels across the dataset, depending on feature space. In contrast, KPMS and Mapper reuse type-level state-pair labels across clips. We retain this asymmetry because it reflects the intrinsic design of each method, and use adjusted mutual information to correct for label-cardinality differences. We use the per-frame rather than per-instance distribution, so a primitive spanning $k$ frames contributes $k$ rows to the cross-tabulation. We report:

\begin{itemize}
    \item $H(T)$: entropy of the coarsening label;
    \item $I(T;Y)=H(Y)-H(Y\mid T)$: mutual information between the coarsening and transition class;
    \item $H(Y\mid T)$: residual transition entropy, or information not preserved by the coarsening;
    \item $\mathrm{NMI}=I(T;Y)/\sqrt{H(T)H(Y)}$: normalized mutual information;
    \item AMI: adjusted mutual information correcting for random partitions of the same cardinality;
    \item shuffle floor: $I(T;Y)$ after independently permuting $T$ within each clip;
    \item lift over shuffle floor: $I-I_{\mathrm{shuffled}}$, isolating cross-clip transition signal from clip-identity effects.
\end{itemize}

Table~\ref{tab:information-loss} reports the eleven variants that enter this comparison, our six DM variants against the two Keypoint-MoSeq and three Mapper baselines; \S\ref{sec:information} states the ranking. The full-clip control is an ablation of the decomposition step rather than a competing coarse-graining, so it is reported with the other controls in Appendix~\ref{app:controls} and is held out of this analysis and of Figure~\ref{fig:ib-panel}. Two further points are visible only here. The DM primitives retain a large lift over the within-clip shuffle floor (DMC-Int $=0.278$), which isolates the frame-level transition signal the decomposition recovers from mere clip identity. And although both baseline families build transitions the same way, as ordered pairs of sustained states, KPMS-KP carries more than three times the transition information of the weakest Mapper variant ($0.525$ against $0.155$ bits), so a state-pair construction is not automatically transition-informative across feature spaces. This analysis complements the alignment evaluation of \S\ref{sec:alignment}: alignment asks whether discovered clusters overlap ground-truth windows, whereas $I(T;Y)$ asks whether the coarse-graining label itself is transition-informative.

\begin{table}[!htbp]
    \centering
    \caption{\textbf{Information-theoretic coarse-graining metrics on rat transitions.} Each row is one method variant, averaged over three seeds for stochastic baselines and deterministic for DM methods. Columns report the entropy of the coarsening label $H(T)$; mutual information $I(T;Y)$; lift over the within-clip shuffle floor; normalized mutual information (NMI); cardinality-corrected adjusted mutual information (AMI); coarsening efficiency $I(T;Y)/H(T)$; and residual transition entropy $H(Y\mid T)$. \textbf{Bold} = best per column, \underline{underline} = second. Baseline methods have different seeds due to their stochasticity in coarse-graining stage whereas DM decomposition is deterministic.}
    \label{tab:information-loss}
\resizebox{\textwidth}{!}{%
\begin{tabular}{lrrrrrrr}
\toprule
Method & $H(T)$ & $I(T;Y)$ & lift over floor & NMI & AMI & efficiency & $H(Y\mid T)$ \\
\midrule
DMC-Ekp & \underline{1.068} & 0.234 & 0.040 & 0.138 & 0.123 & \textbf{0.219} & 2.441 \\
DMC-Intention & 5.184 & \underline{0.873} & \textbf{0.278} & \underline{0.234} & \underline{0.213} & 0.168 & \underline{1.802} \\
DMC-Qpos & 2.853 & 0.474 & 0.121 & 0.172 & 0.166 & 0.166 & 2.200 \\
DMP-Ekp & 1.674 & 0.363 & 0.089 & 0.171 & 0.163 & \underline{0.217} & 2.312 \\
DMP-Intention & 2.977 & 0.511 & 0.162 & 0.181 & 0.172 & 0.171 & 2.164 \\
DMP-Qpos & 4.918 & \textbf{0.895} & \underline{0.276} & \textbf{0.247} & \textbf{0.230} & 0.182 & \textbf{1.780} \\
KPMS-Int & 2.600{\scriptsize$\,\pm\,$0.081} & 0.286{\scriptsize$\,\pm\,$0.036} & 0.164{\scriptsize$\,\pm\,$0.031} & 0.108{\scriptsize$\,\pm\,$0.012} & 0.108{\scriptsize$\,\pm\,$0.012} & 0.110{\scriptsize$\,\pm\,$0.011} & 2.389{\scriptsize$\,\pm\,$0.036} \\
KPMS-KP & 3.908{\scriptsize$\,\pm\,$0.160} & 0.525{\scriptsize$\,\pm\,$0.011} & 0.225{\scriptsize$\,\pm\,$0.004} & 0.162{\scriptsize$\,\pm\,$0.000} & 0.159{\scriptsize$\,\pm\,$0.001} & 0.135{\scriptsize$\,\pm\,$0.003} & 2.150{\scriptsize$\,\pm\,$0.011} \\
Mapper-EKP & 2.313{\scriptsize$\,\pm\,$0.190} & 0.276{\scriptsize$\,\pm\,$0.012} & 0.084{\scriptsize$\,\pm\,$0.007} & 0.111{\scriptsize$\,\pm\,$0.002} & 0.110{\scriptsize$\,\pm\,$0.002} & 0.120{\scriptsize$\,\pm\,$0.006} & 2.399{\scriptsize$\,\pm\,$0.012} \\
Mapper-Int & \textbf{0.940{\scriptsize$\,\pm\,$0.468}} & 0.155{\scriptsize$\,\pm\,$0.086} & 0.060{\scriptsize$\,\pm\,$0.038} & 0.091{\scriptsize$\,\pm\,$0.036} & 0.080{\scriptsize$\,\pm\,$0.040} & 0.154{\scriptsize$\,\pm\,$0.022} & 2.520{\scriptsize$\,\pm\,$0.086} \\
Mapper-Qpos & 2.622{\scriptsize$\,\pm\,$0.093} & 0.322{\scriptsize$\,\pm\,$0.019} & 0.108{\scriptsize$\,\pm\,$0.011} & 0.122{\scriptsize$\,\pm\,$0.007} & 0.121{\scriptsize$\,\pm\,$0.007} & 0.123{\scriptsize$\,\pm\,$0.008} & 2.353{\scriptsize$\,\pm\,$0.019} \\
\bottomrule
\end{tabular}%
}
\end{table}

\subsection{Detailed Alignment \& Consistency Results}\label{app:tables}

The head-to-head comparison of §\ref{sec:alignment} is Table~\ref{tab:main} in the main text, where each family is pinned to a single variant selected by transition information (DMC-Int, DMP-Qpos, Mapper-Qpos, KPMS-KP), so every column corresponds to one runnable configuration. The tables below give the full per-variant grids behind it, including the sync ratio, which the main table omits for space.

The remaining tables give the full per-direction breakdown across all twelve variants (DMC, DMP, and Mapper on intention, qpos, and ekp; KPMS-Int, KPMS-KP, and the full-clip control Full-Int), from which the fixed-space selections above are drawn: Tables~\ref{tab:dtw-detailed} and~\ref{tab:consistency-detailed} for DTW and consistency, Tables~\ref{tab:warping-dtw-sync} and~\ref{tab:warping-non-diag} for the warping decomposition, and Table~\ref{tab:kid-detailed} for KID. All three alignment metrics live in the same root-stripped qpos space; only the kernel changes.

\paragraph{Reading the warping decomposition.} The point of reporting DTW alongside KID is that if a method's DTW lead came from DTW's flexibility to warp away misalignment, the KID column, which cannot warp, would re-rank the methods. To diagnose \emph{how} a method achieves its DTW score, we decompose the warping path into the DTW-to-frame-sync ratio (higher $=$ less warping rescue) and the non-diagonal fraction (lower $=$ a more diagonal alignment, i.e., the primitive matches the transition timescale rather than relying on time-warping). The two point the same way: a method that aligns to ground truth without heavy warping has a high sync ratio and a low non-diagonal fraction.

\definecolor{bestgreen}{HTML}{2E7D32}
\definecolor{secondyellow}{HTML}{F9A825}
\definecolor{worstred}{HTML}{C62828}
\begin{table}[h]
\centering
\caption{\textbf{DTW distance to ground truth by directed transition ($\downarrow$ lower is better).} \textbf{Bold} = best per row over populated cells; \underline{underline} = second. --- = method did not discover this transition. Direction tags: [L] left, [R] right, [S] straight. $^*$Ours. $^\dagger$ marks cells populated by only a single seed (no $\pm$ available). The two summary rows under the table give the per-column mean and standard deviation of the per-cell $\pm$ (std across 3 random seeds), aggregated over the populated rows.}
\label{tab:dtw-detailed}
\tiny
\setlength{\tabcolsep}{1.5pt}
\renewcommand{\arraystretch}{0.95}
\resizebox{\textwidth}{!}{%
\begin{tabular}{l|cccccccccccc}
\toprule
Transition & \shortstack{DMC$^*$\\{\tiny (Int)}} & \shortstack{DMC$^*$\\{\tiny (Qpos)}} & \shortstack{DMC$^*$\\{\tiny (EKP)}} & \shortstack{DMP$^*$\\{\tiny (Int)}} & \shortstack{DMP$^*$\\{\tiny (Qpos)}} & \shortstack{DMP$^*$\\{\tiny (EKP)}} & \shortstack{Mapper\\{\tiny (Int)}} & \shortstack{Mapper\\{\tiny (Qpos)}} & \shortstack{Mapper\\{\tiny (EKP)}} & \shortstack{KPMS\\{\tiny (Int)}} & \shortstack{KPMS\\{\tiny (KP)}} & \shortstack{Full\\{\tiny (Int)}} \\
\midrule
Immobile$\rightarrow$Rear [S] & 0.934 & 0.913 & 0.941 & --- & \textbf{0.892} & 0.948 & 0.906 & \underline{0.895} & 0.945 & 0.961 & 0.967 & 0.949$^{\dagger}$ \\
Immobile$\rightarrow$Rear [L] & --- & 0.929 & 0.934 & 0.927$^{\dagger}$ & 0.854 & 0.966 & \textbf{0.756} & --- & --- & --- & \underline{0.831} & 0.978 \\
Immobile$\rightarrow$Rear [R] & --- & --- & 0.972 & --- & \underline{0.880$^{\dagger}$} & 1.009 & --- & --- & --- & --- & \textbf{0.848$^{\dagger}$} & 0.989 \\
Immobile$\rightarrow$Turn [L] & 0.809 & 0.792 & --- & 0.805 & \underline{0.788} & --- & 0.952$^{\dagger}$ & 0.801 & 0.810 & 0.872 & \textbf{0.770} & 0.817 \\
Immobile$\rightarrow$Turn [R] & 0.759 & 0.732 & --- & \underline{0.727$^{\dagger}$} & 0.756$^{\dagger}$ & --- & 0.810$^{\dagger}$ & \textbf{0.715$^{\dagger}$} & 0.795 & --- & 0.747 & 0.770 \\
Immobile$\rightarrow$Walk [L] & \textbf{0.759$^{\dagger}$} & --- & --- & --- & --- & --- & \underline{0.767} & --- & --- & --- & --- & --- \\
Rear$\rightarrow$Immobile [S] & 0.863$^{\dagger}$ & --- & 1.151$^{\dagger}$ & 0.930 & --- & 0.919 & \textbf{0.823} & \underline{0.856} & 0.886 & 0.926 & 0.877 & --- \\
Rear$\rightarrow$Immobile [L] & --- & 0.888 & 0.926 & --- & 0.855$^{\dagger}$ & 0.884 & 0.892$^{\dagger}$ & 0.879 & \textbf{0.839} & --- & \underline{0.855} & 0.933 \\
Rear$\rightarrow$Immobile [R] & --- & --- & --- & --- & --- & --- & \textbf{0.823$^{\dagger}$} & --- & \underline{0.852$^{\dagger}$} & --- & --- & --- \\
Rear$\rightarrow$Turn [L] & 0.826 & --- & --- & 0.838 & \textbf{0.778} & 0.872 & \underline{0.788} & 0.818 & --- & 0.868 & 0.918 & --- \\
Rear$\rightarrow$Turn [R] & \textbf{0.915$^{\dagger}$} & --- & --- & --- & --- & --- & --- & --- & --- & --- & --- & --- \\
Rear$\rightarrow$Walk [L] & --- & --- & 0.841$^{\dagger}$ & --- & --- & --- & \underline{0.745$^{\dagger}$} & --- & \textbf{0.703$^{\dagger}$} & --- & --- & --- \\
Turn$\rightarrow$Immobile [L] & 0.740 & \textbf{0.720} & --- & 0.797 & \underline{0.721} & --- & 0.722 & 0.790 & --- & --- & 0.849$^{\dagger}$ & 0.778 \\
Turn$\rightarrow$Rear [L] & \textbf{0.907$^{\dagger}$} & --- & --- & --- & --- & --- & --- & --- & --- & --- & --- & --- \\
Turn$\rightarrow$Walk [L] & \textbf{0.643} & --- & --- & --- & --- & --- & 0.698$^{\dagger}$ & --- & --- & \underline{0.644$^{\dagger}$} & --- & --- \\
Walk$\rightarrow$Immobile [S] & --- & --- & --- & \textbf{0.694$^{\dagger}$} & --- & --- & --- & --- & --- & --- & --- & --- \\
Walk$\rightarrow$Turn [L] & --- & --- & --- & \textbf{0.623$^{\dagger}$} & --- & --- & --- & --- & --- & --- & --- & --- \\
\midrule
\textit{Mean of $\pm$} & 0.007 & 0.038 & 0.024 & 0.038 & 0.016 & 0.025 & 0.016 & 0.025 & 0.016 & 0.043 & 0.019 & 0.024 \\
\textit{Std of $\pm$} & 0.003 & 0.030 & 0.012 & 0.023 & 0.011 & 0.017 & 0.010 & 0.012 & 0.012 & 0.019 & 0.011 & 0.024 \\
\bottomrule
\end{tabular}%
}
\end{table}

\definecolor{bestgreen}{HTML}{2E7D32}
\definecolor{secondyellow}{HTML}{F9A825}
\definecolor{worstred}{HTML}{C62828}
\begin{table}[h]
\centering
\caption{\textbf{Internal cluster consistency by directed transition ($\downarrow$ lower is better).} \textbf{Bold} = best per row over populated cells; \underline{underline} = second. --- = method did not discover this transition. Direction tags: [L] left, [R] right, [S] straight. $^*$Ours. $^\dagger$ marks cells populated by only a single seed (no $\pm$ available). The two summary rows under the table give the per-column mean and standard deviation of the per-cell $\pm$ (std across 3 random seeds), aggregated over the populated rows.}
\label{tab:consistency-detailed}
\tiny
\setlength{\tabcolsep}{1.5pt}
\renewcommand{\arraystretch}{0.95}
\resizebox{\textwidth}{!}{%
\begin{tabular}{l|cccccccccccc}
\toprule
Transition & \shortstack{DMC$^*$\\{\tiny (Int)}} & \shortstack{DMC$^*$\\{\tiny (Qpos)}} & \shortstack{DMC$^*$\\{\tiny (EKP)}} & \shortstack{DMP$^*$\\{\tiny (Int)}} & \shortstack{DMP$^*$\\{\tiny (Qpos)}} & \shortstack{DMP$^*$\\{\tiny (EKP)}} & \shortstack{Mapper\\{\tiny (Int)}} & \shortstack{Mapper\\{\tiny (Qpos)}} & \shortstack{Mapper\\{\tiny (EKP)}} & \shortstack{KPMS\\{\tiny (Int)}} & \shortstack{KPMS\\{\tiny (KP)}} & \shortstack{Full\\{\tiny (Int)}} \\
\midrule
Immobile$\rightarrow$Rear [S] & 0.766 & 0.754 & 0.931 & --- & \textbf{0.632} & 0.780 & 0.999 & \underline{0.714} & 0.771 & 0.928 & 0.936 & 0.779$^{\dagger}$ \\
Immobile$\rightarrow$Rear [L] & --- & 0.813 & \textbf{0.744} & 0.877$^{\dagger}$ & \underline{0.803} & 0.979 & 0.831 & --- & --- & --- & 0.821 & 0.868 \\
Immobile$\rightarrow$Rear [R] & --- & --- & \underline{0.823} & --- & \textbf{0.718$^{\dagger}$} & 0.911 & --- & --- & --- & --- & 0.854$^{\dagger}$ & 0.886 \\
Immobile$\rightarrow$Turn [L] & 0.652 & \underline{0.573} & --- & \textbf{0.552} & 0.724 & --- & 1.120$^{\dagger}$ & 0.690 & 0.704 & 0.749 & 0.686 & 0.788 \\
Immobile$\rightarrow$Turn [R] & 0.703 & \textbf{0.638} & --- & 0.706$^{\dagger}$ & 0.729$^{\dagger}$ & --- & 0.844$^{\dagger}$ & 0.669$^{\dagger}$ & \underline{0.648} & --- & 0.688 & 0.768 \\
Immobile$\rightarrow$Walk [L] & \underline{0.928$^{\dagger}$} & --- & --- & --- & --- & --- & \textbf{0.790} & --- & --- & --- & --- & --- \\
Rear$\rightarrow$Immobile [S] & 0.802$^{\dagger}$ & --- & 0.750$^{\dagger}$ & 0.853 & --- & 0.831 & \textbf{0.665} & 0.765 & \underline{0.695} & 0.710 & 0.816 & --- \\
Rear$\rightarrow$Immobile [L] & --- & 0.850 & 0.867 & --- & 0.659$^{\dagger}$ & 0.716 & \textbf{0.551$^{\dagger}$} & 0.743 & \underline{0.655} & --- & 0.708 & 0.913 \\
Rear$\rightarrow$Immobile [R] & --- & --- & --- & --- & --- & --- & \textbf{0.688$^{\dagger}$} & --- & \underline{0.776$^{\dagger}$} & --- & --- & --- \\
Rear$\rightarrow$Turn [L] & 0.782 & --- & --- & 0.825 & \textbf{0.607} & 0.818 & 0.777 & \underline{0.643} & --- & 0.872 & 0.795 & --- \\
Rear$\rightarrow$Turn [R] & \textbf{0.884$^{\dagger}$} & --- & --- & --- & --- & --- & --- & --- & --- & --- & --- & --- \\
Rear$\rightarrow$Walk [L] & --- & --- & \underline{0.683$^{\dagger}$} & --- & --- & --- & 0.930$^{\dagger}$ & --- & \textbf{0.619$^{\dagger}$} & --- & --- & --- \\
Turn$\rightarrow$Immobile [L] & 0.601 & \textbf{0.492} & --- & 0.643 & 0.585 & --- & \underline{0.546} & 0.590 & --- & --- & 0.659$^{\dagger}$ & 0.577 \\
Turn$\rightarrow$Rear [L] & \textbf{0.908$^{\dagger}$} & --- & --- & --- & --- & --- & --- & --- & --- & --- & --- & --- \\
Turn$\rightarrow$Walk [L] & \textbf{0.582} & --- & --- & --- & --- & --- & 0.722$^{\dagger}$ & --- & --- & \underline{0.704$^{\dagger}$} & --- & --- \\
Walk$\rightarrow$Immobile [S] & --- & --- & --- & \textbf{0.556$^{\dagger}$} & --- & --- & --- & --- & --- & --- & --- & --- \\
Walk$\rightarrow$Turn [L] & --- & --- & --- & \textbf{0.562$^{\dagger}$} & --- & --- & --- & --- & --- & --- & --- & --- \\
\midrule
\textit{Mean of $\pm$} & 0.077 & 0.081 & 0.118 & 0.068 & 0.065 & 0.081 & 0.096 & 0.139 & 0.035 & 0.135 & 0.090 & 0.054 \\
\textit{Std of $\pm$} & 0.044 & 0.038 & 0.061 & 0.034 & 0.044 & 0.029 & 0.119 & 0.053 & 0.028 & 0.140 & 0.039 & 0.029 \\
\bottomrule
\end{tabular}%
}
\end{table}

\definecolor{bestgreen}{HTML}{2E7D32}
\definecolor{secondyellow}{HTML}{F9A825}
\definecolor{worstred}{HTML}{C62828}
\begin{table}[h]
\centering
\caption{\textbf{DTW-to-frame-sync ratio by directed transition ($\uparrow$ higher is better).} \textbf{Bold} = best per row over populated cells; \underline{underline} = second. --- = method did not discover this transition. Direction tags: [L] left, [R] right, [S] straight. $^*$Ours. Higher means less reliance on warping rescue. $^\dagger$ marks cells populated by only a single seed (no $\pm$ available). The two summary rows under the table give the per-column mean and standard deviation of the per-cell $\pm$ (std across 3 random seeds), aggregated over the populated rows.}
\label{tab:warping-dtw-sync}
\tiny
\setlength{\tabcolsep}{1.5pt}
\renewcommand{\arraystretch}{0.95}
\resizebox{\textwidth}{!}{%
\begin{tabular}{l|cccccccccccc}
\toprule
Transition & \shortstack{DMC$^*$\\{\tiny (Int)}} & \shortstack{DMC$^*$\\{\tiny (Qpos)}} & \shortstack{DMC$^*$\\{\tiny (EKP)}} & \shortstack{DMP$^*$\\{\tiny (Int)}} & \shortstack{DMP$^*$\\{\tiny (Qpos)}} & \shortstack{DMP$^*$\\{\tiny (EKP)}} & \shortstack{Mapper\\{\tiny (Int)}} & \shortstack{Mapper\\{\tiny (Qpos)}} & \shortstack{Mapper\\{\tiny (EKP)}} & \shortstack{KPMS\\{\tiny (Int)}} & \shortstack{KPMS\\{\tiny (KP)}} & \shortstack{Full\\{\tiny (Int)}} \\
\midrule
Immobile$\rightarrow$Rear [S] & 0.491 & 0.493 & 0.483 & --- & 0.494 & \underline{0.496} & 0.495 & 0.495 & 0.495 & 0.492 & 0.489 & \textbf{0.497$^{\dagger}$} \\
Immobile$\rightarrow$Rear [L] & --- & 0.473 & 0.464 & \textbf{0.490$^{\dagger}$} & 0.476 & 0.477 & 0.448 & --- & --- & --- & 0.462 & \underline{0.484} \\
Immobile$\rightarrow$Rear [R] & --- & --- & 0.477 & --- & \textbf{0.482$^{\dagger}$} & 0.478 & --- & --- & --- & --- & 0.474$^{\dagger}$ & \underline{0.480} \\
Immobile$\rightarrow$Turn [L] & 0.498 & 0.498 & --- & 0.498 & \underline{0.499} & --- & 0.498$^{\dagger}$ & 0.495 & 0.494 & \textbf{0.499} & 0.497 & 0.495 \\
Immobile$\rightarrow$Turn [R] & \underline{0.500} & 0.499 & --- & 0.500$^{\dagger}$ & \textbf{0.500$^{\dagger}$} & --- & 0.499$^{\dagger}$ & 0.498$^{\dagger}$ & 0.494 & --- & 0.498 & 0.498 \\
Immobile$\rightarrow$Walk [L] & \textbf{0.493$^{\dagger}$} & --- & --- & --- & --- & --- & \underline{0.483} & --- & --- & --- & --- & --- \\
Rear$\rightarrow$Immobile [S] & 0.491$^{\dagger}$ & --- & \underline{0.497$^{\dagger}$} & 0.495 & --- & 0.491 & 0.490 & 0.494 & \textbf{0.498} & 0.493 & 0.484 & --- \\
Rear$\rightarrow$Immobile [L] & --- & 0.481 & 0.475 & --- & \textbf{0.491$^{\dagger}$} & 0.481 & 0.475$^{\dagger}$ & \underline{0.484} & 0.479 & --- & 0.483 & 0.482 \\
Rear$\rightarrow$Immobile [R] & --- & --- & --- & --- & --- & --- & \underline{0.442$^{\dagger}$} & --- & \textbf{0.462$^{\dagger}$} & --- & --- & --- \\
Rear$\rightarrow$Turn [L] & 0.482 & --- & --- & \underline{0.488} & 0.477 & 0.484 & 0.472 & 0.473 & --- & \textbf{0.491} & 0.480 & --- \\
Rear$\rightarrow$Turn [R] & \textbf{0.486$^{\dagger}$} & --- & --- & --- & --- & --- & --- & --- & --- & --- & --- & --- \\
Rear$\rightarrow$Walk [L] & --- & --- & \textbf{0.470$^{\dagger}$} & --- & --- & --- & \underline{0.450$^{\dagger}$} & --- & 0.433$^{\dagger}$ & --- & --- & --- \\
Turn$\rightarrow$Immobile [L] & 0.498 & \textbf{0.499} & --- & \underline{0.499} & 0.495 & --- & 0.488 & 0.489 & --- & --- & 0.495$^{\dagger}$ & 0.494 \\
Turn$\rightarrow$Rear [L] & \textbf{0.462$^{\dagger}$} & --- & --- & --- & --- & --- & --- & --- & --- & --- & --- & --- \\
Turn$\rightarrow$Walk [L] & \textbf{0.492} & --- & --- & --- & --- & --- & \underline{0.485$^{\dagger}$} & --- & --- & 0.480$^{\dagger}$ & --- & --- \\
Walk$\rightarrow$Immobile [S] & --- & --- & --- & \textbf{0.496$^{\dagger}$} & --- & --- & --- & --- & --- & --- & --- & --- \\
Walk$\rightarrow$Turn [L] & --- & --- & --- & \textbf{0.489$^{\dagger}$} & --- & --- & --- & --- & --- & --- & --- & --- \\
\midrule
\textit{Mean of $\pm$} & 0.002 & 0.002 & 0.007 & 0.002 & 0.002 & 0.003 & 0.003 & 0.006 & 0.003 & 0.003 & 0.004 & 0.005 \\
\textit{Std of $\pm$} & 0.001 & 0.002 & 0.004 & 0.002 & 0.002 & 0.002 & 0.003 & 0.004 & 0.002 & 0.003 & 0.003 & 0.005 \\
\bottomrule
\end{tabular}%
}
\end{table}

\definecolor{bestgreen}{HTML}{2E7D32}
\definecolor{secondyellow}{HTML}{F9A825}
\definecolor{worstred}{HTML}{C62828}
\begin{table}[h]
\centering
\caption{\textbf{Non-diagonal fraction of DTW warping path by directed transition ($\downarrow$ lower is better).} \textbf{Bold} = best per row over populated cells; \underline{underline} = second. --- = method did not discover this transition. Direction tags: [L] left, [R] right, [S] straight. $^*$Ours. Lower means the warping path runs closer to the diagonal, so the primitive matches the annotated window at its own timescale rather than being stretched or compressed onto it. $^\dagger$ marks cells populated by only a single seed (no $\pm$ available). The two summary rows under the table give the per-column mean and standard deviation of the per-cell $\pm$ (std across 3 random seeds), aggregated over the populated rows.}
\label{tab:warping-non-diag}
\tiny
\setlength{\tabcolsep}{1.5pt}
\renewcommand{\arraystretch}{0.95}
\resizebox{\textwidth}{!}{%
\begin{tabular}{l|cccccccccccc}
\toprule
Transition & \shortstack{DMC$^*$\\{\tiny (Int)}} & \shortstack{DMC$^*$\\{\tiny (Qpos)}} & \shortstack{DMC$^*$\\{\tiny (EKP)}} & \shortstack{DMP$^*$\\{\tiny (Int)}} & \shortstack{DMP$^*$\\{\tiny (Qpos)}} & \shortstack{DMP$^*$\\{\tiny (EKP)}} & \shortstack{Mapper\\{\tiny (Int)}} & \shortstack{Mapper\\{\tiny (Qpos)}} & \shortstack{Mapper\\{\tiny (EKP)}} & \shortstack{KPMS\\{\tiny (Int)}} & \shortstack{KPMS\\{\tiny (KP)}} & \shortstack{Full\\{\tiny (Int)}} \\
\midrule
Immobile$\rightarrow$Rear [S] & 0.080 & 0.092 & 0.143 & --- & 0.063 & 0.063 & 0.045 & 0.055 & \underline{0.032} & 0.097 & 0.096 & \textbf{0.025$^{\dagger}$} \\
Immobile$\rightarrow$Rear [L] & --- & 0.221 & 0.270 & \textbf{0.072$^{\dagger}$} & \underline{0.128} & 0.148 & 0.273 & --- & --- & --- & 0.245 & 0.131 \\
Immobile$\rightarrow$Rear [R] & --- & --- & 0.201 & --- & \textbf{0.170$^{\dagger}$} & \underline{0.173} & --- & --- & --- & --- & 0.173$^{\dagger}$ & 0.184 \\
Immobile$\rightarrow$Turn [L] & \underline{0.012} & 0.023 & --- & 0.022 & 0.019 & --- & 0.040$^{\dagger}$ & 0.033 & 0.051 & \textbf{0.011} & 0.036 & 0.049 \\
Immobile$\rightarrow$Turn [R] & 0.013 & 0.015 & --- & \underline{0.006$^{\dagger}$} & \textbf{0.005$^{\dagger}$} & --- & 0.007$^{\dagger}$ & 0.019$^{\dagger}$ & 0.028 & --- & 0.025 & 0.026 \\
Immobile$\rightarrow$Walk [L] & \textbf{0.080$^{\dagger}$} & --- & --- & --- & --- & --- & \underline{0.089} & --- & --- & --- & --- & --- \\
Rear$\rightarrow$Immobile [S] & 0.080$^{\dagger}$ & --- & 0.053$^{\dagger}$ & 0.053 & --- & 0.108 & 0.080 & \underline{0.052} & \textbf{0.015} & 0.082 & 0.109 & --- \\
Rear$\rightarrow$Immobile [L] & --- & \underline{0.156} & 0.206 & --- & \textbf{0.111$^{\dagger}$} & 0.195 & 0.252$^{\dagger}$ & 0.178 & 0.190 & --- & 0.165 & 0.175 \\
Rear$\rightarrow$Immobile [R] & --- & --- & --- & --- & --- & --- & \underline{0.323$^{\dagger}$} & --- & \textbf{0.265$^{\dagger}$} & --- & --- & --- \\
Rear$\rightarrow$Turn [L] & 0.142 & --- & --- & \underline{0.099} & 0.169 & 0.136 & 0.160 & 0.189 & --- & \textbf{0.085} & 0.164 & --- \\
Rear$\rightarrow$Turn [R] & \textbf{0.135$^{\dagger}$} & --- & --- & --- & --- & --- & --- & --- & --- & --- & --- & --- \\
Rear$\rightarrow$Walk [L] & --- & --- & \textbf{0.214$^{\dagger}$} & --- & --- & --- & \underline{0.261$^{\dagger}$} & --- & 0.334$^{\dagger}$ & --- & --- & --- \\
Turn$\rightarrow$Immobile [L] & 0.032 & \underline{0.024} & --- & \textbf{0.019} & 0.054 & --- & 0.087 & 0.062 & --- & --- & 0.086$^{\dagger}$ & 0.075 \\
Turn$\rightarrow$Rear [L] & \textbf{0.256$^{\dagger}$} & --- & --- & --- & --- & --- & --- & --- & --- & --- & --- & --- \\
Turn$\rightarrow$Walk [L] & \textbf{0.096} & --- & --- & --- & --- & --- & \underline{0.103$^{\dagger}$} & --- & --- & 0.131$^{\dagger}$ & --- & --- \\
Walk$\rightarrow$Immobile [S] & --- & --- & --- & \textbf{0.061$^{\dagger}$} & --- & --- & --- & --- & --- & --- & --- & --- \\
Walk$\rightarrow$Turn [L] & --- & --- & --- & \textbf{0.098$^{\dagger}$} & --- & --- & --- & --- & --- & --- & --- & --- \\
\midrule
\textit{Mean of $\pm$} & 0.011 & 0.015 & 0.024 & 0.023 & 0.017 & 0.022 & 0.023 & 0.029 & 0.015 & 0.018 & 0.018 & 0.036 \\
\textit{Std of $\pm$} & 0.008 & 0.009 & 0.017 & 0.010 & 0.022 & 0.018 & 0.034 & 0.018 & 0.009 & 0.017 & 0.014 & 0.037 \\
\bottomrule
\end{tabular}%
}
\end{table}

\definecolor{bestgreen}{HTML}{2E7D32}
\definecolor{secondyellow}{HTML}{F9A825}
\definecolor{worstred}{HTML}{C62828}
\begin{table}[h]
\centering
\caption{\textbf{Kernel Inception Distance by directed transition ($\downarrow$ lower is better).} \textbf{Bold} = best per row over populated cells; \underline{underline} = second. --- = method did not discover this transition. Direction tags: [L] left, [R] right, [S] straight. $^*$Ours. $^\dagger$ marks cells populated by only a single seed (no $\pm$ available). The two summary rows under the table give the per-column mean and standard deviation of the per-cell $\pm$ (std across 3 random seeds), aggregated over the populated rows.}
\label{tab:kid-detailed}
\tiny
\setlength{\tabcolsep}{1.5pt}
\renewcommand{\arraystretch}{0.95}
\resizebox{\textwidth}{!}{%
\begin{tabular}{l|cccccccccccc}
\toprule
Transition & \shortstack{DMC$^*$\\{\tiny (Int)}} & \shortstack{DMC$^*$\\{\tiny (Qpos)}} & \shortstack{DMC$^*$\\{\tiny (EKP)}} & \shortstack{DMP$^*$\\{\tiny (Int)}} & \shortstack{DMP$^*$\\{\tiny (Qpos)}} & \shortstack{DMP$^*$\\{\tiny (EKP)}} & \shortstack{Mapper\\{\tiny (Int)}} & \shortstack{Mapper\\{\tiny (Qpos)}} & \shortstack{Mapper\\{\tiny (EKP)}} & \shortstack{KPMS\\{\tiny (Int)}} & \shortstack{KPMS\\{\tiny (KP)}} & \shortstack{Full\\{\tiny (Int)}} \\
\midrule
Immobile$\rightarrow$Rear [S] & 0.032 & 0.019 & 0.012 & --- & 0.024 & 0.016 & \textbf{0.007} & \underline{0.010} & 0.016 & 0.022 & 0.011 & 0.021$^{\dagger}$ \\
Immobile$\rightarrow$Rear [L] & --- & 0.010 & \underline{0.005} & 0.028$^{\dagger}$ & \textbf{0.002} & 0.009 & 0.005 & --- & --- & --- & 0.006 & 0.022 \\
Immobile$\rightarrow$Rear [R] & --- & --- & 0.014 & --- & 0.033$^{\dagger}$ & 0.045 & --- & --- & --- & --- & \textbf{0.010$^{\dagger}$} & \underline{0.011} \\
Immobile$\rightarrow$Turn [L] & 0.009 & 0.011 & --- & 0.015 & 0.003 & --- & 0.018$^{\dagger}$ & 0.006 & 0.011 & 0.009 & \underline{0.002} & \textbf{0.001} \\
Immobile$\rightarrow$Turn [R] & 0.007 & 0.011 & --- & 0.006$^{\dagger}$ & 0.009$^{\dagger}$ & --- & 0.008$^{\dagger}$ & 0.008$^{\dagger}$ & 0.011 & --- & \textbf{0.003} & \underline{0.003} \\
Immobile$\rightarrow$Walk [L] & \textbf{0.003$^{\dagger}$} & --- & --- & --- & --- & --- & \underline{0.013} & --- & --- & --- & --- & --- \\
Rear$\rightarrow$Immobile [S] & 0.011$^{\dagger}$ & --- & 0.111$^{\dagger}$ & 0.013 & --- & 0.008 & 0.018 & \underline{0.008} & 0.017 & \textbf{0.004} & 0.008 & --- \\
Rear$\rightarrow$Immobile [L] & --- & \textbf{0.004} & 0.007 & --- & 0.026$^{\dagger}$ & 0.012 & 0.011$^{\dagger}$ & 0.009 & 0.009 & --- & 0.010 & \underline{0.006} \\
Rear$\rightarrow$Immobile [R] & --- & --- & --- & --- & --- & --- & \underline{0.015$^{\dagger}$} & --- & \textbf{0.008$^{\dagger}$} & --- & --- & --- \\
Rear$\rightarrow$Turn [L] & \underline{0.005} & --- & --- & 0.013 & 0.015 & \textbf{0.004} & 0.006 & 0.008 & --- & 0.011 & 0.013 & --- \\
Rear$\rightarrow$Turn [R] & \textbf{0.017$^{\dagger}$} & --- & --- & --- & --- & --- & --- & --- & --- & --- & --- & --- \\
Rear$\rightarrow$Walk [L] & --- & --- & \underline{0.017$^{\dagger}$} & --- & --- & --- & \textbf{0.014$^{\dagger}$} & --- & 0.031$^{\dagger}$ & --- & --- & --- \\
Turn$\rightarrow$Immobile [L] & \underline{0.005} & 0.010 & --- & 0.015 & 0.008 & --- & 0.008 & 0.010 & --- & --- & 0.013$^{\dagger}$ & \textbf{0.002} \\
Turn$\rightarrow$Rear [L] & \textbf{0.004$^{\dagger}$} & --- & --- & --- & --- & --- & --- & --- & --- & --- & --- & --- \\
Turn$\rightarrow$Walk [L] & \underline{0.004} & --- & --- & --- & --- & --- & 0.020$^{\dagger}$ & --- & --- & \textbf{0.003$^{\dagger}$} & --- & --- \\
Walk$\rightarrow$Immobile [S] & --- & --- & --- & \textbf{0.011$^{\dagger}$} & --- & --- & --- & --- & --- & --- & --- & --- \\
Walk$\rightarrow$Turn [L] & --- & --- & --- & \textbf{0.001$^{\dagger}$} & --- & --- & --- & --- & --- & --- & --- & --- \\
\midrule
\textit{Mean of $\pm$} & 0.003 & 0.005 & 0.005 & 0.010 & 0.003 & 0.002 & 0.003 & 0.003 & 0.006 & 0.006 & 0.002 & 0.003 \\
\textit{Std of $\pm$} & 0.003 & 0.003 & 0.005 & 0.007 & 0.002 & 0.002 & 0.004 & 0.002 & 0.003 & 0.005 & 0.002 & 0.004 \\
\bottomrule
\end{tabular}%
}
\end{table}

\subsection{Relation to Supervised and Large-Model Approaches}\label{app:supervised}

A supervised model, including a large transformer, can classify transitions given labels, and with enough compute would likely do so accurately. Our aim is different and complementary: the DM primitives are obtained without transition labels, which are exactly what such a classifier needs and what is expensive to annotate at frame resolution. The output is also an interpretable geometric object, a corridor or a loop in the feature space, rather than an opaque class posterior, and the extraction is cheap, so the method is usable without large-model infrastructure. The contribution is thus an unsupervised, control-grounded representation of transitions, not a competitor to supervised classification accuracy.

\subsection{Computational Cost}\label{app:cost}

DM-G runs once per clip on a commodity CPU and needs no GPU; among the methods compared, only Keypoint-MoSeq's Gibbs sampler benefits from one. The dominant cost is the sparse weighted Rips construction of the reference implementation~\citep{MageeWang2022}: the pairwise-distance matrix and greedy permutation are $O(n^2 D)$, the all-pairs edge enumeration feeding the $\varepsilon$-sparsifier is $O(n^2)$, and the surviving simplex count $m$ then enters PHAT's persistence reduction~\citep{bauer2017phat} at $O(m^{2.37})$. For the $n\!\approx\!500$-frame clips studied here this is tractable per clip.

\subsection{Hyperparameters}
\subsubsection{Hyperparameters for DM-G Methods}\label{app:dmg-hp}

Table~\ref{tab:dm-hp} lists the DM-G hyperparameters used for each feature space and dataset. All feature spaces share $k$-NN connectivity $k{=}3$ (sensitivity analysis in Appendix~\ref{app:k-sweep}) and edge threshold $\varepsilon{=}0.99$. The persistence threshold $\delta$ varies by feature space: intention embeddings tolerate a larger $\delta$ because the 16D space is compact, while qpos (67D after root stripping) and ekp (60D) require smaller $\delta$ to resolve structure in the sparser high-dimensional point cloud. PCA to 10 dimensions is applied to qpos and ekp before DM-G extraction; intention is used without dimensionality reduction. Takens delay $\tau$ is selected as the first local minimum of the Average Mutual Information computed independently per feature space.

\begin{table}[!htbp]
\centering
\caption{DM-G hyperparameters by feature space and dataset.}
\label{tab:dm-hp}
\small
\begin{tabular}{llcccccc}
\toprule
Dataset & Feature space & Raw dim & PCA dim & $m$ & $\tau$ & $k$ & $\delta$ \\
\midrule
rat & Intention & 16 & --- & 3 & 6 & 3 & 1.0 \\
rat & Qpos & 67 & 10 & 3 & 9 & 3 & 0.2 \\
rat & EKP & 60 & 10 & 3 & 8 & 3 & 0.1 \\
\bottomrule
\end{tabular}
\end{table}

Additional shared parameters: $\varepsilon{=}0.99$ for all configurations. Each DM-cycle and DM-path is mapped back to its frames, which are sorted in time and split wherever consecutive frames are more than $50$ frames apart (segment gap limit); segments with fewer than $30$ skeleton vertices are discarded, so one basis cycle or path can yield zero, one or several primitives. For Wasserstein clustering (§\ref{sec:cluster}): target window length $L{=}100$ and $K{=}25$ clusters. The MDS embedding dimension is selected per variant by within-cluster consistency (Appendix~\ref{app:mds-sweep}); the selected values are $20$ (DMC-Int), $40$ (DMC-Qpos), $5$ (DMC-EKP), $10$ (DMP-Int), $40$ (DMP-Qpos) and $40$ (DMP-EKP). Three random seeds $(0, 1, 2)$ are used for all clustering experiments.

\subsubsection{Selection of the MDS Embedding Dimension}\label{app:mds-sweep}

Step 3 embeds the $W_2$ distance matrix by multidimensional scaling before $K$-means (§\ref{sec:cluster}). The embedding dimension governs how much of the transport geometry survives into the clustering step and is a free parameter of the method. We sweep it over $\{3,5,10,20,40,60\}$ for every primitive type and feature space at three seeds, and select per variant the value that minimises \emph{within-cluster consistency}, the mean pairwise DTW among the members of a cluster. Consistency is computed from the discovered clusters alone and never references the ground-truth labels, so the selection is independent of the alignment metrics reported in §\ref{sec:alignment}. Table~\ref{tab:mds-sweep} gives the full grid: block (a) is the selection criterion, and blocks (b) to (d) report the resulting alignment, distributional fidelity and coverage, which take no part in the choice. Means are taken over the transitions a variant populates at every dimension, so a setting cannot appear better merely by discovering a different subset.

No single dimension is best for all variants: the selected values span $5$ to $40$, each inside the swept range. Selecting instead on DTW would change the choice for three of the six variants (DMC-EKP, DMP-Int and DMP-EKP) but for neither of the two reported in Table~\ref{tab:main}, so the ground-truth-free criterion leaves the head-to-head comparison unchanged. We note that the corresponding hyperparameters of the baselines were not swept in the same way; the baseline settings are those of Appendix~\ref{app:baseline-hp}.

\begin{table}[h]
\centering
\caption{\textbf{Selection of the MDS embedding dimension.} Sweep over $\{3,5,10,20,40,60\}$ at three seeds. The dimension is selected per variant by minimising within-cluster consistency (block a), which uses no ground-truth labels; blocks (b)--(d) take no part in the choice. Best per row in \textbf{bold}. Means are over the transitions a variant populates at every dimension. DMP-Int shares only two such transitions, so its block (c) row is noisy.}
\label{tab:mds-sweep}
\scriptsize
\setlength{\tabcolsep}{3pt}
\renewcommand{\arraystretch}{0.85}
\begin{tabular}{lcccccc}
\toprule
& \multicolumn{6}{c}{MDS embedding dimension} \\
\cmidrule(lr){2-7}
Variant & 3 & 5 & 10 & 20 & 40 & 60 \\
\midrule
\multicolumn{7}{l}{\textit{(a) Within-cluster consistency $\downarrow$ (selection criterion)}} \\
DMC-Int & 0.765 & 0.746 & 0.737 & \textbf{0.698} & 0.747 & 0.718 \\
DMC-Qpos & 0.769 & 0.775 & 0.732 & 0.728 & \textbf{0.726} & 0.754 \\
DMC-EKP & 0.872 & \textbf{0.841} & 0.853 & 0.876 & 0.847 & 0.849 \\
DMP-Int & 0.643 & 0.645 & \textbf{0.597} & 0.638 & 0.641 & 0.648 \\
DMP-Qpos & 0.704 & 0.717 & 0.762 & 0.793 & \textbf{0.691} & 0.739 \\
DMP-EKP & 0.915 & 0.862 & 0.874 & 0.884 & \textbf{0.839} & 0.882 \\
\midrule
\multicolumn{7}{l}{\textit{(b) DTW to ground truth $\downarrow$}} \\
DMC-Int & 0.827 & 0.831 & 0.802 & \textbf{0.796} & 0.806 & 0.801 \\
DMC-Qpos & 0.856 & 0.857 & 0.855 & 0.856 & \textbf{0.851} & 0.854 \\
DMC-EKP & 0.935 & 0.943 & 0.943 & 0.933 & \textbf{0.926} & 0.931 \\
DMP-Int & \textbf{0.770} & 0.771 & 0.801 & 0.773 & 0.801 & 0.791 \\
DMP-Qpos & 0.854 & 0.844 & 0.839 & 0.860 & \textbf{0.828} & 0.831 \\
DMP-EKP & \textbf{0.927} & 0.948 & 0.952 & 0.957 & 0.933 & 0.936 \\
\midrule
\multicolumn{7}{l}{\textit{(c) KID $\downarrow$}} \\
DMC-Int & \textbf{0.010} & 0.017 & 0.010 & 0.010 & 0.011 & 0.011 \\
DMC-Qpos & \textbf{0.009} & 0.009 & 0.010 & 0.012 & 0.011 & 0.009 \\
DMC-EKP & 0.010 & 0.009 & \textbf{0.008} & 0.010 & 0.009 & 0.010 \\
DMP-Int & 0.014 & 0.014 & 0.015 & 0.013 & 0.014 & \textbf{0.004} \\
DMP-Qpos & 0.013 & 0.010 & 0.010 & 0.010 & 0.011 & \textbf{0.010} \\
DMP-EKP & 0.014 & 0.012 & 0.014 & \textbf{0.011} & 0.016 & 0.013 \\
\midrule
\multicolumn{7}{l}{\textit{(d) Transitions discovered (union over 3 seeds)}} \\
DMC-Int & 8.0 & 9.0 & 8.0 & 10.0 & 9.0 & \textbf{11.0} \\
DMC-Qpos & 6.0 & \textbf{7.0} & \textbf{7.0} & \textbf{7.0} & 6.0 & 6.0 \\
DMC-EKP & 4.0 & \textbf{6.0} & \textbf{6.0} & \textbf{6.0} & \textbf{6.0} & \textbf{6.0} \\
DMP-Int & 3.0 & 4.0 & \textbf{8.0} & 7.0 & 5.0 & 7.0 \\
DMP-Qpos & 5.0 & 7.0 & 6.0 & 6.0 & \textbf{8.0} & 6.0 \\
DMP-EKP & \textbf{7.0} & 6.0 & 6.0 & 6.0 & 6.0 & 6.0 \\
\bottomrule
\end{tabular}
\end{table}

\subsubsection{Hyperparameters for Baseline Methods}\label{app:baseline-hp}

The KPMS settings below, including the choice of model type, were selected by reconstruction MSE computed on the fitted data for each feature stream. The state-count sweep of Appendix~\ref{app:scale-sweep} uses a different criterion, for the reason given there. Mapper is not fit to data and has no training objective, so its settings are fixed rather than tuned. For every baseline, transition primitives are the twenty-five most frequent ordered pairs of consecutive sustained labels having at least three instances, matching the cluster budget $K{=}25$ given to DM-WKC. Final configurations are listed below.

Keypoint-MoSeq expects observations of shape (frames, bodyparts, spatial dimensions) and, by default, estimates a per-frame heading from a designated anterior--posterior axis and rotates it out. The two feature streams require opposite treatments of that step, and we set them accordingly.

\paragraph{KPMS (intention).} Switching linear dynamical system (SLDS) with latent dimension 10, 25 states, sticky parameter $\kappa{=}1.35 \times 10^4$, 100 AR-only iterations followed by 300 full-model iterations, minimum sustained duration 25 frames. The intention latent is not spatial: it enters as $(T, k, 1)$, one ``bodypart'' per latent coordinate in a single dimension, for which a heading is undefined. We therefore hold the heading fixed at zero so that only the centroid is removed and no rotation is applied.

\paragraph{KPMS (keypoint).} Autoregressive HMM (AR-HMM) with latent dimension 15, 25 states, sticky parameter $\kappa{=}1.09 \times 10^6$, 100 AR-only iterations, minimum sustained duration 25 frames. The keypoints are allocentric marker positions obtained by forward kinematics, so the animal's heading is a nuisance factor: it spans the full circle across the dataset and varies by roughly $190^\circ$ within a single clip. We therefore let KPMS estimate and remove it, taking the head as the anterior landmark and the two hips as the posterior landmarks. This variant is fit in AR-only mode, so the full-model stage that infers the keypoint centroid, heading, and noise scales is not run.

\paragraph{Mapper.}\label{app:mapper-hp} Mapper~\citep{singh2007topological} is run on the three rat feature spaces used for DM-primitive extraction (intention, qpos, ekp), on the same inputs DM-G receives (qpos and ekp reduced by PCA to 10 dimensions) and with the Takens parameters $(m,\tau)$ that DM-G uses in each space (Table~\ref{tab:dm-hp}). The 1D lens is the Fiedler vector, the eigenvector of the smallest non-zero eigenvalue of the normalised Laplacian of an unweighted $k$-nearest-neighbour graph over the pooled cloud, computed with \texttt{scikit-learn}'s spectral embedding. We start from $k{=}15$ rather than DM-G's $k{=}3$, since a $k$-NN \emph{connectivity} graph needs a larger $k$ than a $k$-NN \emph{density} estimate and at $k{=}3$ the Takens-embedded filaments break into components, on which spectral embedding silently returns a component indicator instead of a Fiedler vector. We therefore test connectivity and double $k$ until it holds: intention and ekp connect at $k{=}15$, while qpos needs $k{=}30$ on one seed and $k{=}60$ on the other two, the growth being seed-dependent because the graph is fit on a random subsample. The raw lens is heavy-tailed, placing over $99.8\%$ of intention points in one decile on every seed and so collapsing the cover to a single interval, so we rank-normalise it, a monotone reparameterisation that preserves its level sets; the arbitrary eigenvector sign is pinned deterministically. All clips are pooled into one cloud carrying one graph, which is what makes cross-clip node pairs meaningful; for tractability it is fit on a uniform random subsample of $10^{5}$ of the roughly $4\times10^{5}$ points, the rest inheriting their nearest subsample neighbour's assignment. The cover is the standard cubical cover with $n_{\text{intervals}}{=}10$ and overlap fraction $0.5$; inner clustering is DBSCAN with $\texttt{min\_samples}{=}5$ and one global neighbourhood radius, the median pairwise distance of a $10^{3}$-point sample. Overlaps are resolved by assigning each frame to the containing node whose cover interval is centred nearest its lens value, and primitives are then formed as for KPMS: runs of at least 50 frames with one node label count as sustained, each cluster is one ordered pair of consecutive sustained labels, and the twenty-five most frequent pairs with at least three instances are retained, leaving Takens warm-up and DBSCAN noise frames unlabelled. This configuration differs from the Mapper variant swept in Appendix~\ref{app:scale-sweep}.

\subsection{Reproducibility}\label{app:repro}

All code, ground-truth labels, and extracted primitives needed to reproduce the results will be released accordingly, containing: (i) the full analysis pipeline (Takens embedding, DM-G skeletonization, DM-cycle and DM-path extraction, and the $W_2\!\to$MDS$\to K$-means clustering of §\ref{sec:cluster}); (ii) the heuristic auto-labeling and proofreading pipeline of Appendix~\ref{app:labeling} together with the proofread \texttt{rat\_labels.csv}; and (iii) configuration files that generate every table and figure in the paper.

\paragraph{Determinism and seeds.} DM-G skeletonization and primitive extraction are deterministic given a feature space and its hyperparameters (Appendix~\ref{app:dmg-hp}), so the seed spread of the DM-WKC results comes only from the MDS embedding and the $K$-means stage. The stochastic baselines and the MDS and $K$-means stages are run over three fixed seeds $(0, 1, 2)$, and the tables report their mean and standard deviation. Mapper's only stochastic elements are the subsample used for the graph fit and the sample used to set the DBSCAN neighbourhood radius (Appendix~\ref{app:mapper-hp}), so its seed spread reflects subsampling rather than the algorithm itself. The subsample also fixes the neighbourhood size $k$ at which the lens graph first becomes connected, so on qpos that value varies across seeds.

\paragraph{Inputs.} The trajectories analyzed are rollouts of a trained MIMIC-MJX imitation policy~\citep{mimicmjx2025}. The derived per-clip feature arrays (intention, qpos, ekp) that DM-G consumes were extracted from those rollouts and will be released with the code, so the analysis can be reproduced without access to the emulator itself. The upstream markerless-capture recordings remain subject to the terms of their original release~\citep{mimicmjx2025}.

\paragraph{Hyperparameters and entry points.} All DM-G and baseline hyperparameters are listed in Appendices~\ref{app:dmg-hp} and~\ref{app:baseline-hp}, and the repository README maps each reported table and figure to the exact command that produces it.

\subsection{Discovered Cluster Examples}\label{app:cluster-examples}

\begin{figure}[!htbp]
    \centering
    \includegraphics[width=0.7\textwidth]{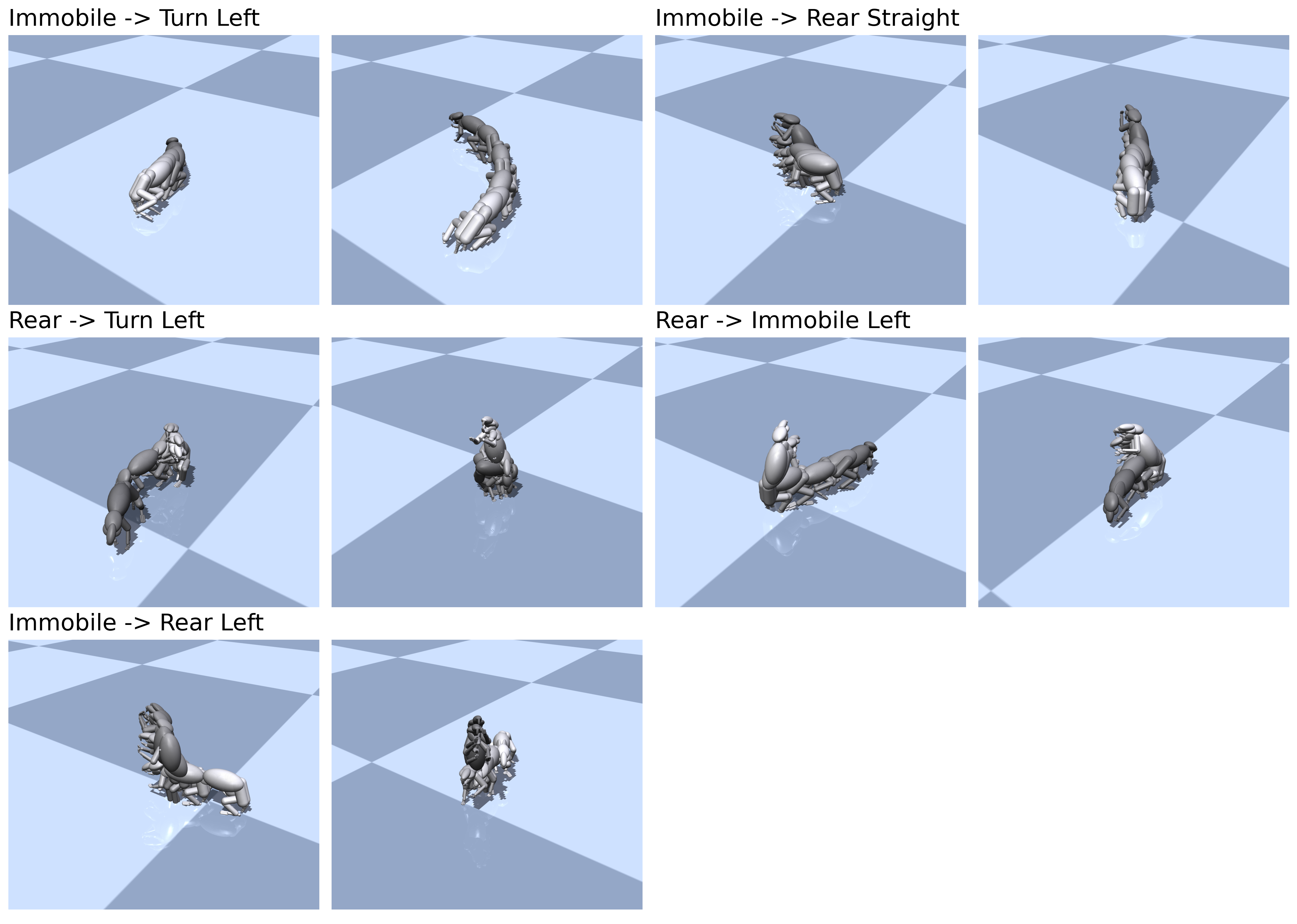}
    \caption{\textbf{DMP-Qpos cluster examples}, seed 0. Two clusters per row, two members per cluster, six poses per member superimposed in-scene under a shared camera and tinted light-to-dark by time. The title above each pair is the cluster's majority-vote label. All five of this variant's unique labels are shown.}
    \label{fig:examples-dmp-qpos}
\end{figure}

\begin{figure}[!htbp]
    \centering
    \includegraphics[width=0.7\textwidth]{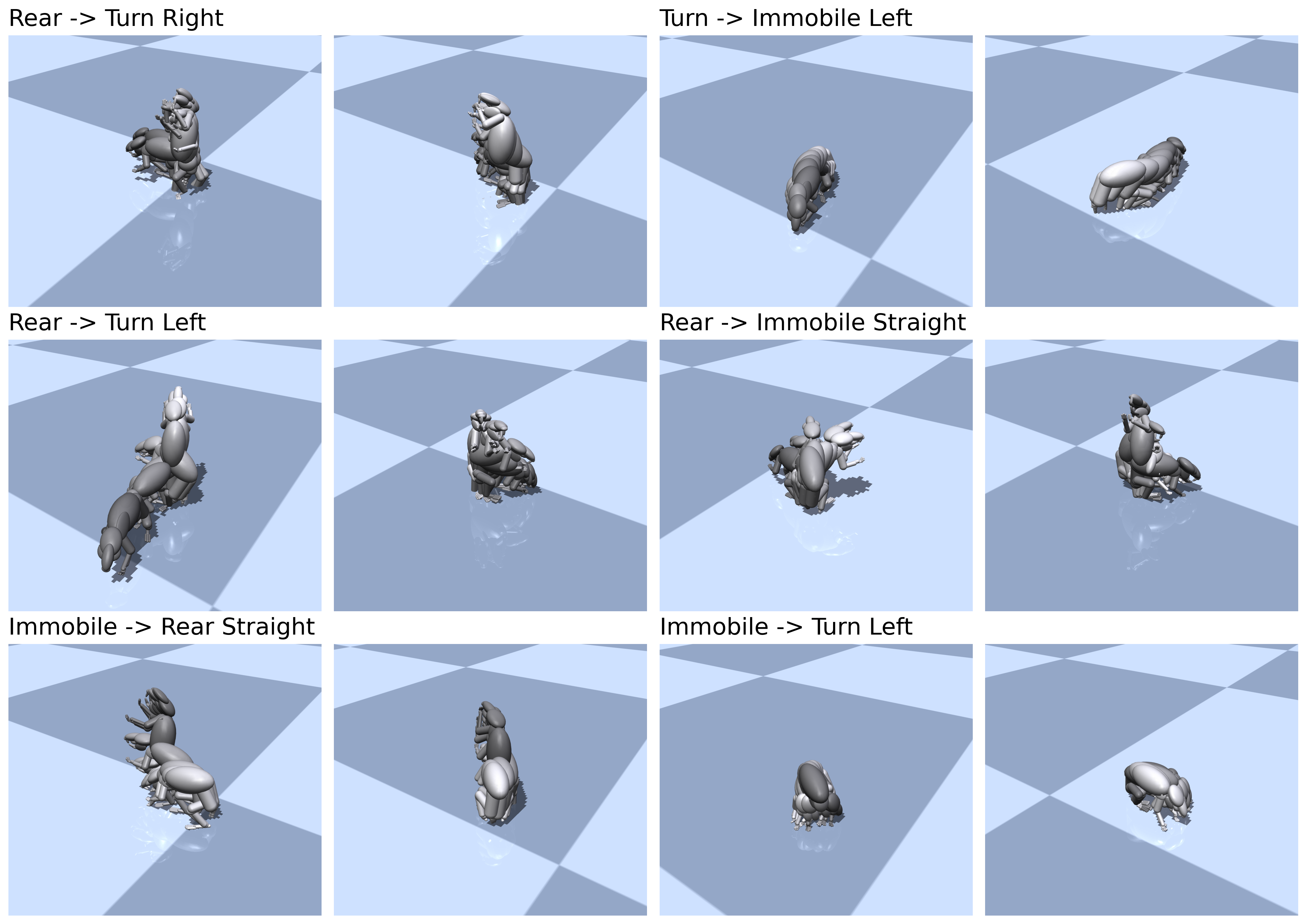}
    \caption{\textbf{DMC-Int cluster examples.} Same layout as Figure~\ref{fig:examples-dmp-qpos}; all six of this variant's unique labels are shown.}
    \label{fig:examples-dmc-int}
\end{figure}

\begin{figure}[!htbp]
    \centering
    \includegraphics[width=0.7\textwidth]{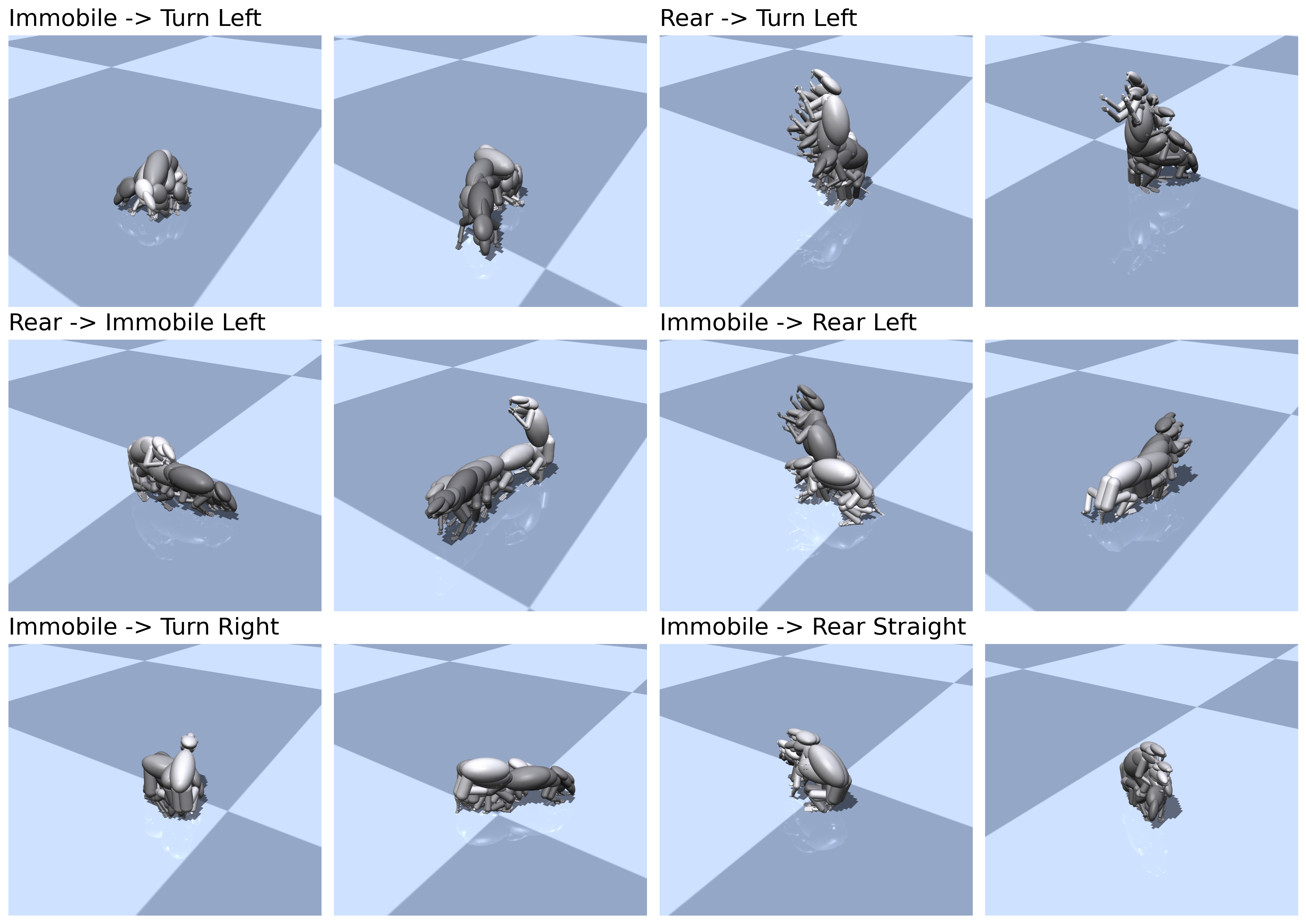}
    \caption{\textbf{KPMS-KP cluster examples.} Same layout as Figure~\ref{fig:examples-dmp-qpos}. KPMS-KP yields 7 unique cluster labels at this seed, against 5 for DMP-Qpos and 6 for DMC-Int; six of the seven are shown, to keep every panel of this appendix on one page.}
    \label{fig:examples-kpms-kp}
\end{figure}

\begin{figure}[!ht]
    \centering
    \includegraphics[width=0.7\textwidth]{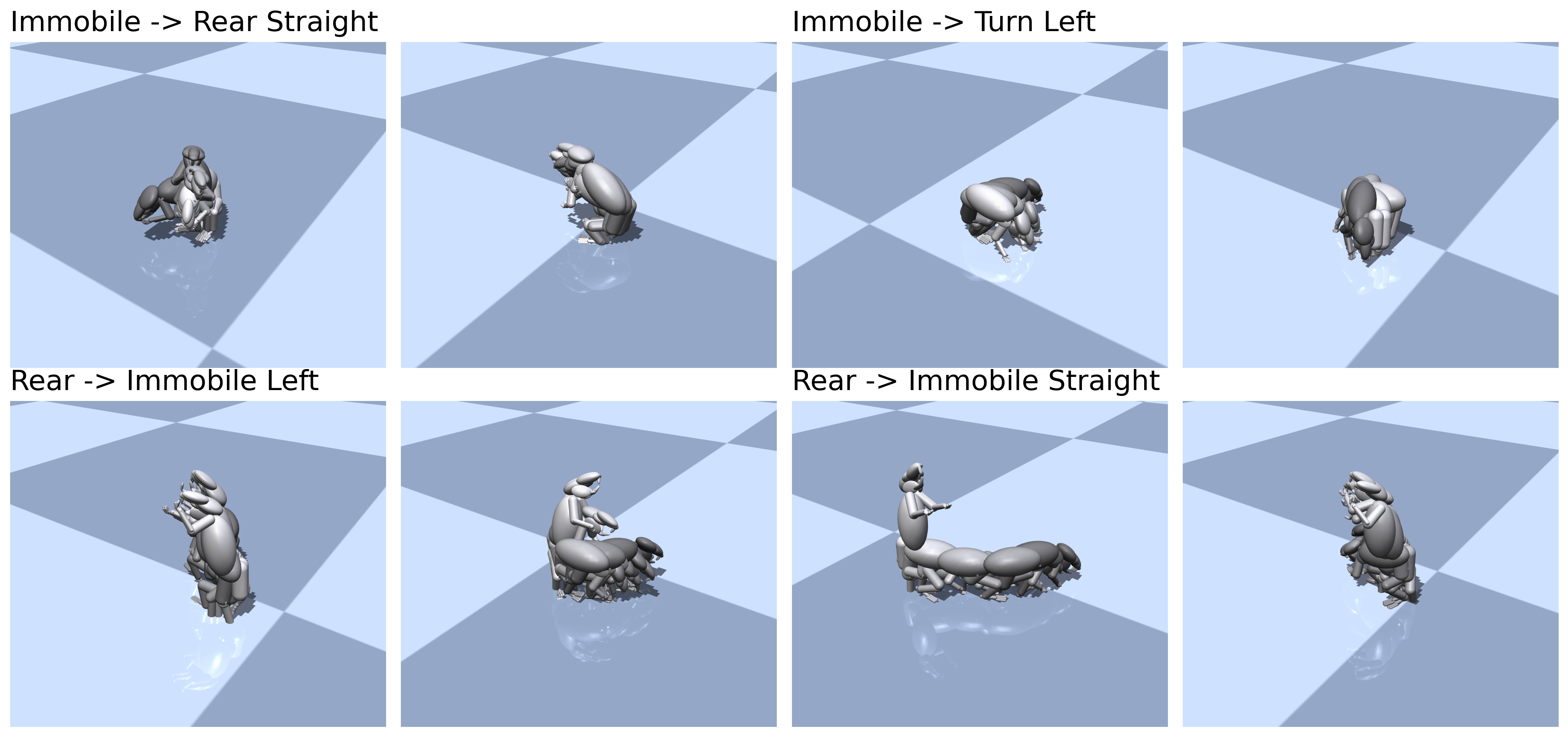}
    \caption{\textbf{Mapper-Qpos cluster examples.} Same layout as Figure~\ref{fig:examples-dmp-qpos}, for the Mapper variant reported in Table~\ref{tab:main}; all four of its unique labels are shown.}
    \label{fig:examples-mapper-ekp}
\end{figure}

Figures~\ref{fig:examples-dmp-qpos},~\ref{fig:examples-dmc-int},~\ref{fig:examples-kpms-kp} and~\ref{fig:examples-mapper-ekp} render discovered clusters for the four variants reported in Table~\ref{tab:main}: DMP-Qpos, DMC-Int, and the two baselines at their reported variants, KPMS-KP and Mapper-Qpos. Each figure shows up to six unique cluster labels (transition type with turn direction), two clusters per row and two members per cluster. Members are superimposed in the scene rather than laid out as a strip, so the poses sit where the animal actually was and the shape of the trajectory is visible alongside the posture. Within each cluster the two members should trace the same gross movement, which is the visual counterpart of the coherence numbers in Table~\ref{tab:main}.

\end{document}